\documentclass[11pt,a4paper]{article}
\usepackage{times,latexsym}
\usepackage{url}
\usepackage[T1]{fontenc}

\usepackage[utf8]{inputenc} 
\usepackage[T1]{fontenc}    
\usepackage{hyperref}       
\usepackage{url}            
\usepackage{booktabs}       
\usepackage{amsfonts}       
\usepackage{nicefrac}       
\usepackage{microtype}      
\usepackage{xcolor}         
\usepackage{microtype}
\usepackage{graphicx}
\usepackage{hyperref}
\usepackage{wrapfig}
\usepackage{soul}

\usepackage[english]{babel}

\usepackage{times}
\usepackage{latexsym}
\usepackage{paralist}
\usepackage{xspace}
\usepackage[normalem]{ulem}
\useunder{\uline}{\ul}{}
\usepackage{tabu}
\usepackage{tabularx}
\usepackage{makecell}
\usepackage{multirow}
\usepackage{ulem}
\usepackage[small,compact]{titlesec}
\usepackage{amsmath}
\usepackage{amssymb}
\usepackage{mathtools}
\usepackage{amsthm}

\usepackage{lipsum}
\usepackage{titlesec}
\usepackage{setspace}

\usepackage{titlesec}
\usepackage[capitalize,noabbrev]{cleveref}

\theoremstyle{plain}
\newtheorem*{definition}{Definition}
\theoremstyle{definition}

\usepackage[textsize=tiny]{todonotes}

\usepackage{times}
\usepackage{latexsym}
\usepackage{paralist}
\usepackage{xspace}
\usepackage[normalem]{ulem}
\useunder{\uline}{\ul}{}
\usepackage{tabu}
\usepackage{tabularx}
\usepackage{makecell}
\usepackage{multirow}
\usepackage{ulem}
\usepackage{algorithm}
\usepackage{algorithmic}
\usepackage{multicol}

\usepackage{amsmath,amsfonts,bm}

\def\eqref#1{equation~\ref{#1}}

\def\1{\bm{1}}

\def\vs{{\bm{s}}}

\DeclareMathAlphabet{\mathsfit}{\encodingdefault}{\sfdefault}{m}{sl}
\SetMathAlphabet{\mathsfit}{bold}{\encodingdefault}{\sfdefault}{bx}{n}

\newcommand{\EE}{\mathop{\mathbb{E}}}

\newcommand{\R}{\mathbb{R}}

\DeclareMathOperator*{\argmin}{arg\,min}

\makeatletter
\DeclareRobustCommand\onedot{\futurelet\@let@token\@onedot}
\def\@onedot{\ifx\@let@token.\else.\null\fi\xspace}

\def\posthoc{\textit{post-hoc}\xspace}
 
\def\eg{\textit{e.g}\onedot} 
\def\ie{\textit{i.e}\onedot} 
 
 \def\vs{\textit{vs}\onedot}
\def\wrt{\textit{w.r.t}\onedot} 

\makeatother

\usepackage{color, colortbl}
\definecolor{emerald}{rgb}{0.31, 0.78, 0.47}
\definecolor{Gray}{gray}{0.9}
\definecolor{Highlight}{rgb}{0.89,0.89,0.94}
\usepackage[first=0,last=9]{lcg}
\newcommand{\chl}{\cellcolor{Highlight}}

\usepackage{titlesec}
\titlespacing{\paragraph}{0pt}{\parskip}{\parskip}
\titlespacing{\section}{0pt}{\parskip}{\parskip}
\titlespacing{\subsection}{0pt}{\parskip}{\parskip}
\titlespacing{\subsubsection}{0pt}{\parskip}{\parskip}

\renewcommand{\bm}[1]{\mathbf{#1}}

\newcommand{\z}[1]{\bm{z}_{#1}}

\newcommand{\emb}[1]{\textsc{Emb}(#1)}

\newcommand{\method}{\textsc{{DiNoiser}}\xspace}

\newcommand{\revise}[1]{#1}

\def\code#1{\texttt{#1}}

\newcommand{\textbi}[1]{\textbf{\textit{#1}}}
\newcommand{\smallcitep}[1]{{\footnotesize\citep{#1}}}
\usepackage{stackengine,scalerel}
\newcommand\dddag{%
  \sbox0{\ddag}\scalerel*{%
  \stackengine{-.6\ht0}{\ddag}{\ddag}{O}{c}{F}{F}{S}}{\ddag}%
} 
\usepackage[acceptedWithA]{tacl2021v1}

\usepackage{xspace,mfirstuc,tabulary}

\newif\iftaclinstructions
\taclinstructionsfalse 
\iftaclinstructions
\renewcommand{\confidential}{}
\renewcommand{\anonsubtext}{(No author info supplied here, for consistency with
TACL-submission anonymization requirements)}
\newcommand{\instr}
\fi

\iftaclpubformat 

\else

\fi

\title{\method: Diffused Conditional Sequence \\Learning by Manipulating Noises}

\author{
  Jiasheng Ye\Thanks{This work was done during Jiasheng's internship at ByteDance Research. $^\dagger$~Zaixiang Zheng is the corresponding author.}~~$^{\heartsuit \diamondsuit}$ ~ Zaixiang Zheng$^{\dagger\heartsuit}$ ~ Yu Bao$^\heartsuit$ ~ Lihua Qian$^\heartsuit$ \and Mingxuan Wang$^\heartsuit$
  \\
  \ 
  $^\heartsuit$ByteDance Research~~$^\diamondsuit$Fudan University
  \\
  \texttt{\small jsye23@m.fudan.edu.cn,~zhengzaixiang@bytedance.com} \\
  \texttt{\small \{baoyu.001, qianlihua, wangmingxuan.89\}@bytedance.com} \\
  \texttt{\small \href{https://github.com/yegcjs/DINOISER}{\textcolor{magenta}{https://github.com/yegcjs/DINOISER}} }
}

\date{}

\begin{document}
\maketitle
\begin{abstract}
While diffusion models have achieved great success in generating continuous signals such as images and audio, it remains elusive for them to learn discrete sequence data like natural languages.
Although recent advances circumvent this challenge of discreteness by embedding discrete tokens as continuous surrogates, they still fall short of satisfactory generation quality.
To understand this, we first dive deep into the denoised training protocol of diffusion-based sequence generative models and determine their three severe problems: (1) failing to learn; (2) lack of scalability; and (3) neglecting source conditions. 
We argue that these problems can be boiled down to the \textit{pitfall of the not completely eliminated discreteness} in the embedding space, and the \textit{scale of noises }is decisive herein. 
In this paper, we introduce \method to facilitate diffusion models for sequence generation by manipulating noises.
We propose to adaptively determine the range of sampled noise scales during training; and encourage the proposed diffused sequence learner to leverage source conditions with amplified noise scales during inference.
Experiments show that \method enables consistent improvement over the baselines of previous diffusion sequence generative models on several conditional sequence modeling benchmarks thanks to both effective training and inference strategies. 
Analyses further verify that \method can make better use of source conditions to govern its generative process.
\end{abstract}

\section{Introduction}
Conditional sequence learning aims at generating a target sequence from given conditions, which is one of the important paradigms of natural language generation~\citep{sutskever2014seq2seq,wiseman2017data2text,raffel2020T5}, including machine translation~\citep{bahdanau2014neural}, summarization~\citep{rush-etal-2015-neural}, and paraphrasing~\citep{DBLP:journals/corr/PrakashHLDQLF16}.
Recent advances in generative modeling introduce diffusion models~\citep{diffusion2015,ddpm,song2020sde}, which achieve great success in generating continuous signals, including images~\citep{rombach2021highresolution}, video~\citep{ho2022imagenvideo}, and audio~\citep{kong2020diffwave}. 
\revise{With promising characteristics such as diversity and controllability demonstrated in these domains, diffusion models also garner growing interest for sequence learning in the research community~\citep{diffusionlm}, which further gives the promise to a unified generative modeling paradigm across different modalities~\citep{bao2023one}, }

\begin{figure*}[t]
    \centering
    \includegraphics[width=\linewidth]{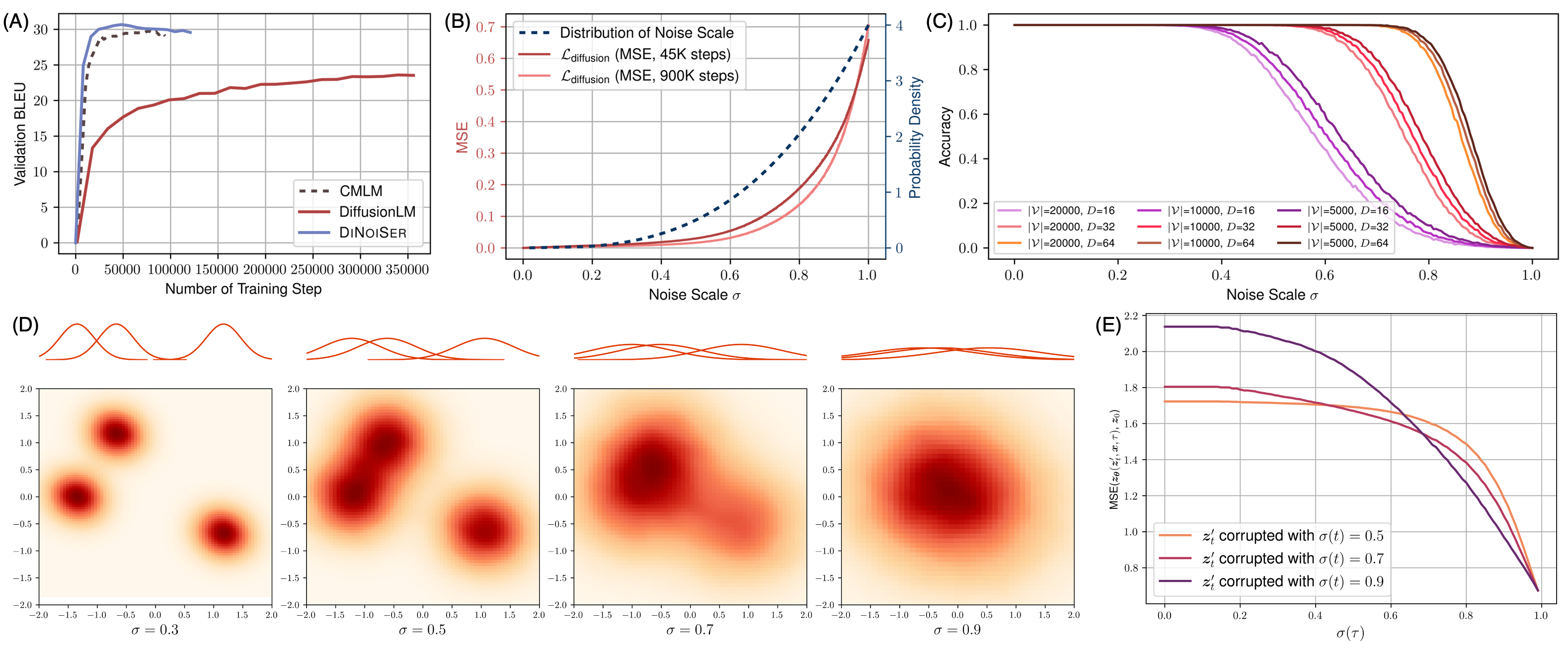}
    \vspace{-6mm}
    \caption{\textit{Preliminary study.}
    \textbf{(A)} 
    The validation BLEU of different models on IWSLT14 \textsc{De}$\rightarrow$\textsc{En} at different training steps. 
    \textbf{(B)} Diffusion loss of DiffusionLM on the validation set of IWSLT14 \textsc{De}$\rightarrow$\textsc{En} at different noise scales and the distribution of noise scale sampled during training.
    \textbf{(C)} The accuracy of predicting $\bm{z}_0$ from $\bm{z}_t$ by finding the nearest neighbor for $\bm{z}_t$ with different noise scales, vocabulary sizes $|\mathcal{V}|$, and dimensions $D$. 
    \textbf{(D)} An illustrative example of the distributions of $\z{t}$ of three data points corrupted with different noise scales as in Eqn.~\ref{eqn:diffusion}, where for small noise scales, a large proportion of the embedding space between modes (associated with tokens) remains vacant.
    \textbf{(E)} The tendency of whether the model prediction is more influenced by the source or target side information when fed with timestep correspond to different noise scales.
    In addition to the source condition $\bm{x}$, we feed the model with $\bm{z}_t^\prime = \z{t}(\bm{y}^{\prime}, t)$ that is corrupted with a timestep-dependent noise $\sigma(t)$ from a negative $\bm{y}^\prime$, which is different from the original (positive sample of) target sequence $\bm{y}$.
    We compare the similarity between the model prediction $\bm{z}_\theta(\bm{z}_t^\prime, \bm{x}, \tau)$ to the embedding of ground-truth $\z{0}(\bm{y})$, and study to what degree the model prediction is governed by the source condition $\bm{x}$ (via the embedding of the ground-truth $\bm{z}_0(\bm{y})$ as the proxy), or the target information $\bm{y}^{\prime}$ (via $\bm{z}_t^\prime$) otherwise.
    }
    \label{fig:prelim}
\end{figure*}

However, the discrete nature of sequence data, constituted by a number of tokens in order, makes it non-trivial to apply diffusion models for conditional sequence learning.
Typical diffusion models noise data with Gaussian permutation kernels~\citep{ddpm} and learn to recover original data from their corrupted versions, which is not directly compatible with discrete tokens.
To remedy this, DiffusionLM~\citep{diffusionlm} attempted to embed discrete tokens into continuous space and employ diffusion models to the embedding space. 
Although this kind of approach unlocks the possibility of applying diffusion models to discrete data, it still falls short of competitive performance for various conditional sequence generation tasks (Fig.~\ref{fig:prelim}A).

We argue that embedding discrete tokens into continuous surrogates does not necessarily eliminate discreteness completely.
To verify this, we conduct in-depth preliminary studies and highlight our findings along with their implications as follows.
\textbf{(1)}~On the \textit{pitfall of discreteness}. 
Embeddings populate only finite clusters~(up to the vocabulary size) in the continuous space, which results in the vastness of low-density regions especially when the models are learned with small-scale noises. 
We refer to this as the pitfall of discreteness, which suggests that small noises hinder conditional sequence learning, and thus should be avoided during training.
\textbf{(2)}~On \textit{scalability}. 
It becomes increasingly harder for the diffusion process to eliminate discreteness when the dimension of the embedding space gets scaled up, suggesting that to ensure scalability, an adaptable noise schedule is necessitated yet neglected.
\textbf{(3)}~On \textit{conditional learning}. 
Enlarging noises in inference can calibrate diffusion models to take into account more source conditional information.
Please refer to \S\ref{sec:preliminary} for more details.

Motivated by these findings, we propose \method to improve \textbf{\underline{di}}ffusion models by manipulating \textbf{\underline{noi}}ses for conditional \textbf{\underline{se}}quence lea\textbf{\underline{r}}ning.
We propose a novel training strategy to eliminate training on small noise scales to avoid their negative influences, for which we introduce the noise scale clipping strategy to adaptively manipulate the noise scales.
For inference, we propose to manipulate the model to be exposed to larger noise scales to encourage trained diffusion models to leverage source conditions.

We summarize our contributions as well as our findings as follows:
\begin{compactitem}
    \item By thorough and in-depth preliminary studies, we shed light on the pitfall of discreteness along with the critical role of noise scales in conditional sequence learning with diffusion models, thereby suggesting meliorated solutions in terms of both training and inference by manipulating noises.
    
    \item We accordingly propose \method to leverage large noise scales in both training and inference. 
    Experiments show that \method achieves strong performance on a variety of conditional sequence learning tasks, paving way for featuring diffusion models for various conditional sequence learning tasks. Our experiments comprehensively include several machine translation benchmarks (both bilingual and multilingual), as well as text simplification and paraphrasing, ranging from low-resource to high-resource scenarios.
    
    \item Ablations show that both \method’s improved training and inference approaches result in considerable performance gains. Further analysis verifies that our proposed posthoc inference strategy, i.e., the condition enhanced denoiser, can help make better use of source conditions for accurate predictions.
\end{compactitem}

\section{Background}

\paragraph{Conditional Sequence Learning.}
Conditional sequence learning aims to yield target sequence $\bm{y}=[y_1, y_2, \dots, y_n] \in \{0,1\}^{n \times |\mathcal{V}|}$ within the vocabulary space $\mathcal{V}$, given source conditions $\bm{x}$, which can be another sequence $\bm{x}=[x_1, x_2, \dots, x_m]$. 
The conventional modeling paradigm~\citep{sutskever2014seq2seq,vaswani2017attention} generates target tokens in an autoregressive decomposition $p(\bm{y}|\bm{x}) = \prod_{i=1}^n p(y_i| \bm{y}_{<i},\bm{x})$.
\citet{gu2018non}~proposed an alternative way in a fully non-autoregressive~(NAR) manner, where all the tokens are predicted in parallel by assuming conditional independence between the target tokens, \ie, $p(\bm{y}|\bm{x}) = \prod_{i=1}^n p(y_i|\bm{x})$.
Later works alleviate this strong assumption by iterative refinement~\citep{lee2018deterministic,ghazvininejad2019mask,gu2019levenshtein}, resulting in improved generation quality. 
These iterative refinement approaches generate target sequences with several cycles, in each of which the models generate sequence depending on both the source sequence and the intermediate prediction of the previous one, \ie, $p(\bm{y}|\bm{x})=\prod_{t=1}^T p(\bm{y}^{(t)}|\bm{y}^{(t-1)}, \bm{x})$.

\paragraph{Diffusion Probabilistic Models.}
Given a random variable $\bm{z}_0$ from an underlying data distribution $q(\bm{z}_0)$, diffusion models~\citep{diffusion2015,ddpm} define a forward diffusion process $\{\bm{z}_t\}_{t\in[0, 1]}$ perturbed with a Gaussian perturbation kernel, starting with $\bm{z}_0$ and converging to its corresponding stationary distribution, such that for any $t\in[0, 1]$, the distribution of $\bm{z}_t$ given $\bm{z}_0$ satisfies 
$
q(\bm{z}_t|\bm{z}_0) = \mathcal{N}(\bm{z}_t ; \alpha(t)\bm{z}_0, \sigma^2(t)\bm{I})$, or with Gaussian reparameterization~\citep{kingma2013auto,ddpm},
\begin{equation}
     \bm{z}_t = \alpha(t)\z{0} + \sigma(t)\bm{\epsilon}_t, ~~\bm{\epsilon}_t \sim \mathcal{N}(\bm{0}, \bm{I}),
     \label{eqn:diffusion}
\end{equation}
where $\sigma(t)$ is a monotonically increasing function, usually referred to as the noise schedule, satisfying $\sigma(0)=0$ and $\sigma(1)\approx 1$;
and $\alpha(t)=\sqrt{1-\sigma^2(t)}$.
The noise schedule~$\sigma(t)$ controls the degree of corruption at different timestep~$t$. 
As $t$ gets larger, the noise scale $\sigma(t)$ gets larger whereas $\alpha(t)$ gets smaller, hence the more corrupted data $\z{t}$ from the original $\z{0}$.
At $t=1$, with $\alpha(1)\approx 0$ and $\sigma(1)\approx 1$, $\bm{z}_t$ become pure noises as reaching the stationary distribution of a standard Gaussian.

\citet{song2020sde} proves that such a Gaussian diffusion process has the same transition distribution $q(\bm{z}_t|\bm{z}_0)$ as the stochastic differential equation (SDE):
$\mathrm{d}\bm{z} = -\frac{1}{2}\beta(t)\bm{z} \mathrm{d}t + \sqrt{\beta(t)}\mathrm{d}\bm{\omega}$
where $\beta(t)=-2\frac{\mathrm{d}\log \alpha(t)}{\mathrm{d}t}$;
$\bm{\omega}$ denotes the standard Wiener process.
As such, the corresponding generative process can be achieved as its time reversal by solving the following ordinary differential equation (diffusion ODE): 
\begin{equation}
\begin{aligned}
\!\!  \mathrm{d}\bm{z} &= \left[-\frac{1}{2}\beta(t)\bm{z}  + \frac{1}{2} \beta(t) 
  \frac{\bm{\epsilon}_t}{\sigma(t)}
  \right] \mathrm{d}t \\
  &= \left[-\frac{1}{2}\beta(t)\bm{z}  + \frac{\beta(t)}{2\sigma^2(t)} \left(\bm{z}-\alpha(t)\z{0}\right)\right] \mathrm{d}t.
  \label{eqn:diffode}
\end{aligned}
\end{equation}

In practice, we can then use a learned model $\bm{z}_{\bm{\theta}}(\z{t}, t)$ to estimate $\z{0}$ and plug into Eqn.~\ref{eqn:diffode}, which can be learned by minimizing the discrepancy between training data and model estimation \citep{ddpm,song2020sde,ramesh2022dalle2}:
\begin{equation}
\begin{aligned}
\!\!\mathcal{L_{\text{diffusion}}}(\z{0}) = \! \EE_{\substack{t \sim \mathcal{U}(0,1) \\ \bm{\epsilon}_t \sim \mathcal{N}(\bm{0}, \bm{I})}} \! \Big[ \|\bm{z}_{\bm{\theta}}(\bm{z}_t, t) - \bm{z}_0\|_2^2 \Big].
\label{eqn:loss_diffusion}
\end{aligned}
\end{equation}
Given Eqn. \ref{eqn:diffode} with a trained model $\bm{z}_{\bm{\theta}}(\z{t}, t)$, we can use arbitrary ODE solvers to solve this diffusion ODE from $t=1$ to $t=0$ for sampling data.
An effective and efficient solver to this end is the DDIM solver~\citep{ddim,lu2022dpm} and is widely adopted.
It discretizes the ODE into $M+1$ timesteps $\{t_i\}_{i=0}^M$ decreasing from $t_0=1$ to $t_M \approx 0$. 
Then, it samples $\z{t_0}$ from the standard Gaussian distribution and computes $\{\bm{z}_{t_i}\}_{i=1}^M$ with $M$ iterations, in each of which $\bm{z}_{t_{i}}$ is predicted from $\bm{z}_{t_{i-1}}$ according to 
\begin{equation}
\small
\resizebox{0.9\columnwidth}{!}{$
    \bm{z}_{t_{i}} = \alpha(t_i) \bm{z}_{\bm{\theta}} (\bm{z}_{t_{i-1}}, t_{i-1}) + \sigma(t_{i}) \bm{\epsilon}_{\bm{\theta}}(\z{t_{i-1}}, t_{i-1}),
$}
\label{eqn:ddim}
\end{equation}
where $\bm{\epsilon}_{\bm{\theta}}(\z{t_{i-1}}, t_{i-1})$ is the predicted noise, which can be directly induced according to Eqn.~\ref{eqn:diffusion},
\begin{equation}
\small
\resizebox{0.9\columnwidth}{!}{$
    \bm{\epsilon}_{\bm{\theta}}(\z{t_{i-1}}, t_{i-1}) = \frac{\bm{z}_{t_{i-1}}- \alpha(t_{i-1})\bm{z}_{\bm{\theta}} (\bm{z}_{t_{i-1}}, t_{i-1}) }{\sigma(t_{i-1})}.
$}
\label{eqn:predicted-noise}
\end{equation}
After iterations, the last prediction $\bm{z}_{t_M}$ is taken as the final generated result $\bm{\hat{z}}_{0}$ of the sampling.

\paragraph{Diffusion Models for Conditional Sequence Learning.}
\label{sec:diffusion4clm}
The denoising process of diffusion models matches an iterative refinement process~\citep{gong2022diffuseq}. 
However, diffusion models are not directly applicable to sequence learning tasks since the original diffusion models operate in continuous space rather than sequences of discrete tokens. 
DiffusionLM~\citep{diffusionlm} tackles this by embedding the discrete tokens into continuous latent space and applying diffusion models therein.
We can then train the models as variational autoencoders~\citep{kingma2013auto}, where a diffusion model serves as the prior, from a latent-variable model perspective, and derive the corresponding variational lower bound~\citep{diffusionvae,latentsde}:
\begin{equation}
\small
    \mathcal{L}(\bm{y}) = \mathbb{E}_{\z{0}} \Big[ \underbrace{-\log p_{\bm{\theta}}(\bm{y}|\bm{z}_0)}_{\mathcal{L}_{\text{reconstruction}}} 
    + \mathcal{L}_{\text{diffusion}}(\z{0}) \Big], 
\label{eqn:difflm_obj}
\end{equation}
where $\bm{y}$ is the original sequence with $\z{0}$ as its embeddings\footnote{DiffusionLM adds tiny noise to the embeddings to form $\z{0}$ (\ie $\z{0}\sim\mathcal{N}(\emb{\bm{y}}, \sigma_0\bm{I})$). We empirically find this unnecessary and letting $\z{0}$ follow a Dirac distribution makes training more efficient.};
$\mathcal{L}_{\text{diffusion}}$ denote the diffusion loss (Eqn.~\ref{eqn:loss_diffusion}) which now operates on the embedding space, and $\mathcal{L}_{\text{reconstruction}}$ is the newly added reconstruction term.  

To further adapt the model for conditional sequence generation, a vanilla approach is to replace the unconditional model $\bm{z_\theta}(\bm{z}_t, t)$ with a conditional model $\bm{z_\theta}(\bm{z}_t, \bm{x}, t)$, where $\bm{x}$ is the source condition. 
Similar to the previous practice of using diffusion models for conditional generation in vision~\citep{rombach2021highresolution}, the diffusion process can be kept unchanged, the same as Eqn.~\ref{eqn:diffusion}. 
And the length of the target sequences is decided by predicting the length difference between the source and the target.


\section{The Pitfall of Discreteness: The Noise Scale Matters}
\label{sec:preliminary}
In this section, we dive deep into the current weaknesses of diffusion models for conditional sequence learning and find that the noise scale matters, which accordingly motivates our proposal for improved training and inference. 
\paragraph{Settings.}
We begin with the vanilla conditional diffusion model modified from DiffusionLM~\citep{diffusionlm} as described in \S\ref{sec:diffusion4clm}.
We follow the original paper of DiffusionLM to apply the \texttt{sqrt} schedule (\ie, $\sigma(t) = t^{0.25}$) to arrange noise scales for training and sampling.
We use IWSLT14 \textsc{De}$\rightarrow$\textsc{En}~\citep{cettolo2012wit3} machine translation benchmark for evaluation.
We also include CMLM~\citep{ghazvininejad2019mask} as a baseline for comparison, which is a strong conditional sequence generative model that generates sequence by iterative refinement similar to diffusion models but in discrete tokens space.

\paragraph{Observations.} 
Here are our findings.

\begin{compactitem}
\item[O1.] \textbi{DiffusionLM still falls short of conditional sequence learning.}
Fig.~\ref{fig:prelim}(A) shows the validation performance of the two models at different training steps, in which the performance of DiffusionLM still lags behind CMLM by a large margin, even taking many more steps before convergence. 
This shows that the performance and training efficiency of the vanilla diffusion-based sequence learner remain unsatisfactory.

\item[O2.] \textbi{Diffusion losses at small noise scales are unexpectedly small.}
DiffusionLM uniformly samples timesteps hence the corresponding noise scales during training.
As shown in Fig.~\ref{fig:prelim}(B), we find that the magnitudes of diffusion losses approach almost zero for small noise scales, indicating that it is quite trivial to recover the corrupted embeddings under such circumstances.
We conjecture that, combined with the illustrated example in Fig.~\ref{fig:prelim}(D), this is because there remain highly discrete modes for the embedding density such that any corrupted embedding is very likely to lie in a region with a small radius around the original token embedding.
As a consequence, the more the modes of embeddings separate from each other the smaller the diffusion loss,  which adheres to the following observation.

\item[O3.] \textbi{It becomes increasingly harder for the diffusion process to eliminate discreteness while the dimension of the embedding space scales up.}
Fig.~\ref{fig:prelim}(C) shows a surprisingly high accuracy of recovering corrupted embeddings that can be easily achieved by simply seeking the nearest neighbor when embedding dimensions enlarge, even at considerably large noise scales.
This reveals that scaling embedding space leads to more severe discreteness, namely a curse of dimensionality.


\item[O4.] \textbi{On condition learning: larger noise scales calibrate diffusion models in taking into account more source conditional information during inference.}
\label{sec:problem2}
We have seen that recovering embeddings corrupted with small noise scales is easy~(O2), and if modes distribute separately, even finding the nearest neighbor is enough (O3). 
In Fig.~\ref{fig:prelim}(C), as the noise scale decreases, the prediction accuracy by finding the nearest neighbor increases and achieves almost 100\% under a threshold, which can be learned trivially even with little to no source conditions.
This results in the hallucination as shown in Tab.~\ref{tab:hallucination}, which is an unexpected consequence for conditional sequence generative models to yield output loyal to the input condition. 
To mitigate this, we quantitatively study the influence of noise scales on conditional reliance.
As shown in Fig~\ref{fig:prelim}(E), we find that as the noise scales are larger, the model can predict more faithfully to source conditions. 
\end{compactitem}


\begin{table}[t]
\centering
\small
\caption{Illustration of hallucinations of vanilla DiffusionLM on IWSLT14 \textsc{De}$\rightarrow$\textsc{En} translation task, along with \method's predictions, where the vanilla DiffusionLM generates inexplicable expressions that are irrelevant to the source condition whose meaning dramatically differs from the groud-truth target.
}
\label{tab:hallucination}
\resizebox{\columnwidth}{!}{
\begin{tabular}{ll}
\toprule
\textbf{Source} & Mit welchen worten würden sie ban beschreiben? \\
\textbf{Reference} & What are the words you would use to \textcolor{red}{describe} ban? \\
\midrule
\textbf{DiffusionLM} & In which words would you \textcolor{red}{save} ban? \\
\textbf{\method} & In which words would you \textcolor{red}{describe} ban? \\ 
\bottomrule
\end{tabular}
}
\end{table}

\paragraph{Concluding remarks.}
We summarize conclusions from the aforementioned observations along with suggestions for more plausible diffused conditional sequence learning:
\begin{compactitem}
    \item[C1.] \textbi{We should not train on too small noise scales to circumvent the pitfall of discreteness.} 
    Both O2 and O4 show the negative influences of small noise scales on training that it leads to a not smooth embedding space with vast regions of low density between modes associated with tokens (O2). 
    These regions can inevitably be sampled during inference\footnote{Consider that a token $\alpha$ is translated into token $A$ or $a$ with 50\% each. 
    Without extra information, the optimal prediction for translating $\alpha$ is the center of the embedding of $A$ and $a$. 
    This is because minimizing its training objective (\ie, $\frac{1}{2} \| \bm{z}_{\bm{\theta}} - \bm{z}_{A}(0)\|_2^2  + \frac{1}{2}\| \bm{z}_{\bm{\theta}} - \bm{z}_{a}(0)\|_2^2 $) results in $\bm{z}_{\bm{\theta}}=\frac{\bm{z}_{A}(0) + \bm{z}_{a}(0)}{2}$.
    The prediction exactly falls in the blank area that lies between embeddings. }, thereby giving rise to error accumulation.
    Besides, it also impedes conditional learning (O4).
    To remedy this, probably a simple way is to eliminate the chance of training with small noise scales.
    
    \item[C2.] \textbi{We need to determine the noise schedule according to the dimensionality of the embedding space.}
    Fitting more complex datasets usually requires larger embedding dimensions. 
    O3 indicates the criterion to distinguish large and small noise scales depends on the embeddings hence the complexity of the datasets.
    However, existing methods employ a fixed noise schedule for all embedding dimensions, which lacks scalability.
    This, therefore, demands a task-specific noise schedule to accommodate diverse datasets.

    \item[C3.] \textbi{We could expose the model to larger noise scales for better source conditions leverage. }
    O4 suggests that the more corrupted the embeddings, the more difficult for the model to recover, thereby necessitating more reliance on source conditions.
    Accordingly, we may encourage trained diffusion models to care more about source conditions for free by \posthoc manipulating the noise to large ones. 
    
\end{compactitem}

\section{\method}

Provided the observations and postulates we discussed in \S\ref{sec:preliminary}, we accordingly propose \method, a simple yet effective method that improves \underline{\textbf{di}}ffusion models by manipulating \underline{\textbf{noi}}ses for conditional \underline{\textbf{se}}quence lea\underline{\textbf{r}}ning. 
The general principle of \method is to determine the best-suited noise scales for both training and inference for conditional sequence generation.
In a nutshell, as for training, we propose to eliminate the chance of training diffused sequence learners with small-scale noises so as to circumvent the aforementioned pitfall of discreteness in embedding space (\S\ref{sec:clipping}).
As for sampling, we propose a new effective sampler to amplify the impact of source conditions on the model prediction, where timesteps corresponding to large noise scales are always fed into the model (\S\ref{subsec: sampling}).
We now dive deep into the details of \method.

\subsection{Noise Scale Clipping: Manipulating Noises for Counter-Discreteness Training }
\label{sec:clipping}

\begin{figure}[t]
    \vspace{-2mm}
    \centering
    \includegraphics[width=\linewidth]{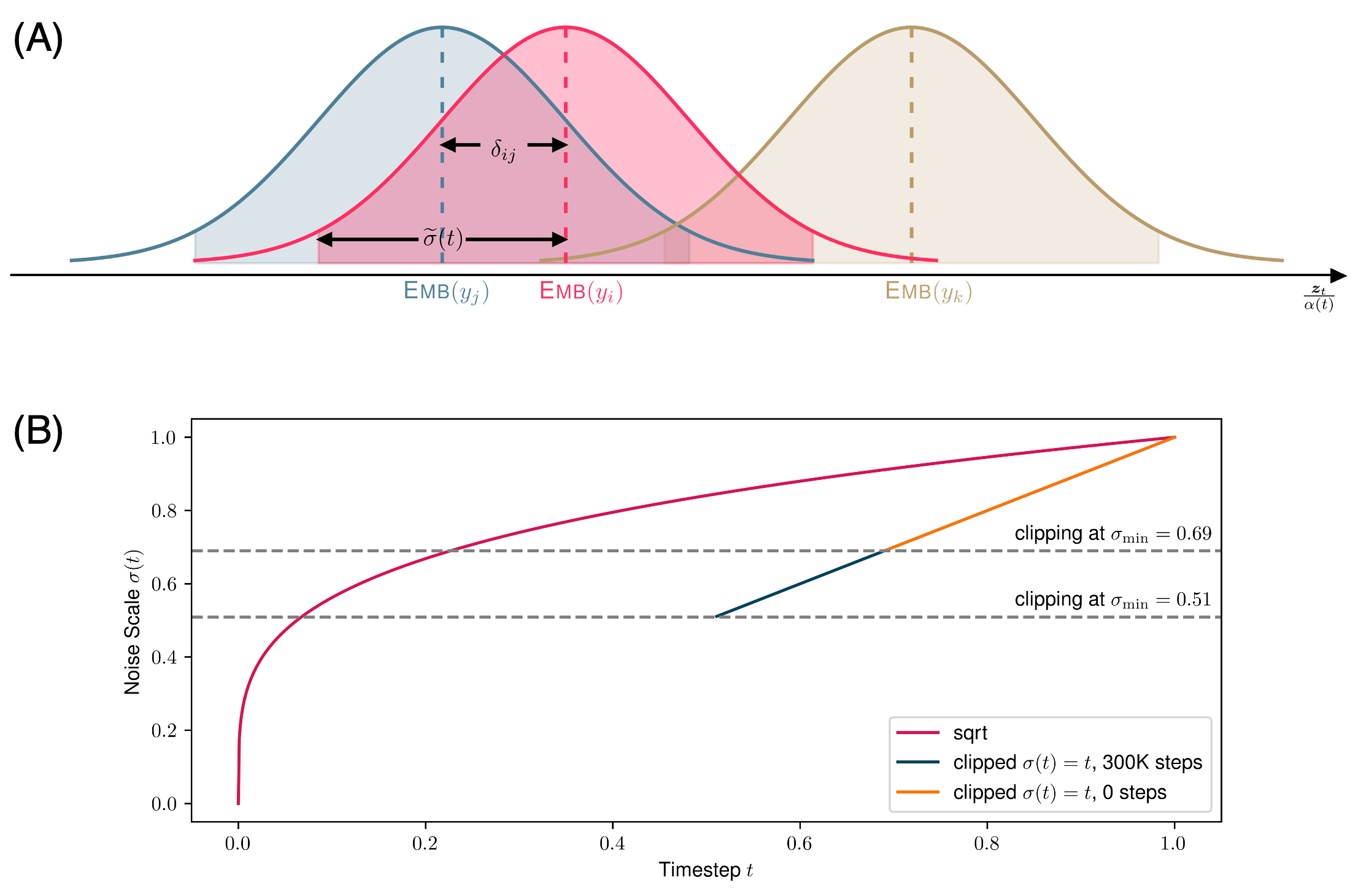}
    \vspace{-7mm}
    \caption{\textbf{(A)}~Illustration of the proposed noise scale clipping.
    To remedy the pitfall of discreteness, we propose to ensure a sufficiently large minimum ``overlap'' between corrupted embeddings.
    As shown in this example, such a goal of counter-discreteness is achieved by bounding the standard deviation $\Tilde{\sigma}(t)$ of $\emb{y_i}$ by $\delta_{ij}$ the ``distance'' to its nearest neighbor (\ie, $y_j$).
    \textbf{(B)}~Comparison between \texttt{sqrt} noise schedule and our noise schedule $\sigma(t)=t$ manipulated with the proposed noise scale clipping.}
    \label{fig:noise-clipping}
    \vspace{-4mm}
\end{figure}

Recall that C1 and C2 in \S\ref{sec:preliminary} demonstrate that small noises can barely help ``discrete'' embeddings populate the entire continuous space, and also undermine conditional learning. 
A simple yet effective way to mitigate this is to encourage training diffusion models with sufficiently large noise scales.
As such, we propose \textit{noise scale clipping}, where we bound the minimum noise scale $\sigma_{\min}$ for training such that only timesteps satisfying $\sigma(t)\ge\sigma_{\min}$ could be sampled, which is decided adaptively as the model learn progresses.

To start with, we can eliminate the scaling effect of $\alpha(t)$ in the forward diffusion process for each token embedding by rewriting Eqn.~\ref{eqn:diffusion} into:
\begin{equation}\small
\resizebox{0.9\columnwidth}{!}{$
\begin{aligned}
\frac{\bm{z}_t[i]}{\alpha(t)} = \bm{z}_0 &+ \frac{\sigma(t)}{\alpha(t)}\bm{\epsilon}_t \Rightarrow \frac{\bm{z}_t[i]}{\alpha(t)} \sim \mathcal{N}\biggl(\bm{z}_0[i], \frac{\sigma^2(t)}{1-\sigma^2(t)}\bm{I} \biggr)  \\
&\Rightarrow \frac{\bm{z}_t[i]}{\alpha(t)} \sim \mathcal{N}\biggl(\emb{y_i}, \Tilde{\sigma}^2(t)\bm{I} \biggr)  
\end{aligned}
$}
\end{equation}
As illustrated in Fig.~\ref{fig:noise-clipping}(A), there, intuitively, should exist a sufficiently large number $\delta^*$ measuring the minimum ``overlap'' between the distributions of two corrupted embeddings under the Gaussian perturbation kernel with a standard deviation of $\Tilde{\sigma}(t) = \frac{\sigma(t)}{\sqrt{1-\sigma^2(t)}}$.
\revise{To this end, we let $\delta^2$ be the minimum amount of variation of added noise, defined as the average squared L2-distances between the embeddings and their nearest neighbor, normalized by the dimension of embeddings (according to C2 in \S\ref{sec:preliminary}): 
\begin{equation}
\small%
\resizebox{0.9\columnwidth}{!}{$
\begin{aligned}
    (\delta^*)^{2} &=\frac{1}{{|\mathcal{V}|}}\sum_{i=1}^{|\mathcal{V}|} \min_{1\leq j\not=i \leq |\mathcal{V}|} \delta^2_{ij} \\
    &=\frac{1}{{|\mathcal{V}|}}\sum_{i=1}^{|\mathcal{V}|} \min_{1\leq j\not=i \leq |\mathcal{V}|} \frac{1}{D} \|\emb{y_i}-\emb{y_j}\|_2^2.
\end{aligned}
$}
\label{eqn:min-dist}
\end{equation}
}
We now define the noise scale clipping\footnote{\revise{\textbf{Our goal can be motivated through the lens of optimal transport}.
That it to say, we aim to determine the minimum cumulative cost $\delta^2 = \sum_{i=1}^{L} \mathbf{T}_{ij} \delta^2_{ij}$, where $L$ is the sequence length, by finding the optimal transportation $\mathbf{T}$ of moving the perturbed embeddings at timestep $t$, \ie $\frac{\bm{z}_t[i]}{\alpha(t)} \sim \mathcal{N}\biggl(\emb{y_i}, \Tilde{\sigma}^2(t)\bm{I} \biggr),~~ i \in \mathcal{V} $, such that the corrupted embedding $\frac{\bm{z}_t[i]}{\alpha(t)}$, if gets noised by a Gaussian deviation} \revise{satisfying $\Tilde{\sigma}(t) >= \delta^*$, cannot be discriminated from those originate from different embeddings, otherwise a smaller $\Tilde{\sigma}(t)$ will lead to trivial reconstruction to the original one as a consequence of the pitfall of discreteness (O2 \& O3 in \S\ref{sec:problem2}).
As a result, $\delta^*$ serves as a minimum clipping threshold of noise scale for effective training of sequence diffusion models.}\\
\revise{This can closely relate to the minimum Word Mover Distance, a Wasserstein metric introduced in \citet{kusner2015word}: 
\begin{equation}
\begin{aligned}
    \delta^2 = \min_{\mathbf{T}} \sum_{i=1}^{L} \mathbf{T}_{ij} \delta^2_{ij},~~~~ \mathrm{s.t.}~ \sum_j \mathbf{T}_{ij} = d_i, \nonumber
\end{aligned}
\end{equation}
where $\mathbf{T} \in \R^{L \times L}$ is a (sparse) stochastic matrix, where $\mathbf{T}_{ij}$ denotes \textit{how much} of a word $i$ travels to word $j$, subject to the flow consistency equality $\sum_j \mathbf{T}_{ij} = d_i$, with $d_i$ representing the ``amount'' of a word $i$ appearing in token embedding space (we treat $d_i = 1$).
According to the Eqn. (2) in \citet{kusner2015word}, under mild conditions, the optimal solution $\mathbf{T}^*$  is for each token $i$ to move all its
probability mass to the most similar token $j$ \wrt a certain measure of their embedding distances, 
$\delta_{ij} = \|\frac{\bm{z}_t[i]}{\alpha(t)}-\emb{y_j}\|_2$:
\begin{equation}
    \mathbf{T}^*_{ij} = \begin{cases}
        d_i &\text{if}~j=\argmin_{1\leq j\not=i \leq L} \delta^2_{ij} \\
        0 &\text{otherwise}
    \end{cases}. \nonumber
\end{equation}
As a result, the final minimum transportation cost becomes:
\begin{align}
    (\delta^*)^{2} & = \sum_{i=1}^{L} \mathbf{T}^*_{ij} \delta^2_{ij} 
     = \sum_{i=1}^{L} d_i \cdot (\delta^2_{ij})^* =  \sum_{i=1}^{L} \min_{1\leq j \not=i \leq L} \delta^2_{ij} \nonumber \\
     & = \sum_{i=1}^{L}  \min_{1\leq j \not=i \leq L}  \Bigl[\left\|\frac{\bm{z}_t[i]}{\alpha(t)}-\emb{y_j}\right\|_2\Bigr]^2\nonumber
\end{align}
A too-small noise scales result in that the nearest neighbors of the corrupted embeddings are exactly their origins, thus
\begin{align}
\small
    (\delta^*)^{2}&=
    &=\sum_{i=1}^L  \left\|\frac{\bm{z}_t[i]}{\alpha(t)} - \emb{y_i}\right\|_2^2=\sum_{i=1}^L\left(\Tilde{\sigma}(t)\bm{\epsilon_i}\right)^2,
    \nonumber
\end{align}
where $\bm{\epsilon_i}$ are standard Gaussian noises.
The above results contain no model parameters, indicating that a diffusion model, which learns to minimize the Wasserstein distance between prediction and target distribution~\citep{kwon2022score}, can not learn from those mildly perturbed samples. 
From this perspective, our noise clipping tries to avoid training on these unhelpful samples.}
} as follows:
\begin{definition}[The noise scale clipping]
Let $\mathcal{V}$ be the target vocabulary with corresponding embeddings in $D$-dimensional space $\forall y_i \in \mathcal{V}: \emb{y_i} \in \R^{D}$, the noise scale clipping is performed so that the noise scale $\sigma(t)$ always satisfies:
\begin{equation}
\small
\resizebox{0.9\columnwidth}{!}{$
    ~~~~~\Tilde{\sigma}^2(t) = \frac{\sigma^2(t)}{1-\sigma^2(t)} \geq (\delta^*)^2 ~\Rightarrow~ \frac{\sigma^2_{\min}}{1-\sigma^2_{\min}} = (\delta^*)^2,
\label{eqn:trunc}
$}
\end{equation}
the clipping threshold $\sigma_{\min}$ is whereby derived when the equality in Eqn.~\ref{eqn:trunc} holds, such that
\begin{equation}\small%
\resizebox{0.85\columnwidth}{!}{$
   \!\!\!\! \sigma_{\min} \!= \! \Biggl(\frac{{|\mathcal{V}|}\cdot D}{\sum_{i=1}^{|\mathcal{V}|} \mathop{\min}\limits_{1\leq j\not=i \leq |\mathcal{V}|} \|\emb{y_i}-\emb{y_j}\|_2^2} + 1 \! \Biggr)^{\!\!-\frac{1}{2}}\!\!\!\!\!\!,
$}
\label{eqn:sigma-min}  
\end{equation}
which is obtained by substituting Eqn.~\ref{eqn:min-dist} into the R.H.S of Eqn.~\ref{eqn:trunc}.
\end{definition}

As illustrated in Fig.~\ref{fig:noise-clipping}(B), the clipping threshold $\sigma_{\min}$ is estimated \textit{adaptively} during training, depending on how properly the model learns the embeddings up to the minimum pair-wise distances within the vocabulary. 
In each training step, we first estimate the clipping threshold $\sigma_{\min}$ with Eqn.~\ref{eqn:sigma-min}, then sample timesteps among $t$ that satisfies $\sigma(t)>\sigma_{\min}$.
In practice, one can first estimate the noise scale threshold $\sigma_{\min}$ and then turn it into the timestep threshold $t_{\min} = \sigma^{-1}(\sigma_{\min})$ in general. 
In this work, we select $\sigma(t)=t$ as the noise scheduler to simplify this procedure\footnote{This can be done since the effects of different noise schedules are theoretically interchangeable up to different weight factors under the simplified training objective~\citep{ddpm} we adopted (see Appendix \ref{sec:appendix scheduler}). We also provide empirical comparisons between different schedules in Tab.~\ref{tab:ablation}. 
}.

\begin{algorithm}[t]\small
\caption{Training with \method} 
\label{alg:training}
\textbf{Input} Training dataset $\mathcal{D}=\{(\bm{x}, \bm{y})\}$.\\
\textbf{Output} Optimized parameters $\bm{\theta}$.

\begin{algorithmic}[1]
  \REPEAT
  \STATE Sample $\bm{x}, \bm{y}$ from the dataset $\mathcal{D}$ and embed $\bm{y}$ into $\z{0}$

  \STATE $t\sim\mathcal{U}(\sigma^{-1}(\textcolor{red}{\sigma_{\text{min}}}),1)$, where $\sigma_{\min}$ is from Eqn.~\ref{eqn:sigma-min}
  \STATE Sample $\z{t}$ with Gaussian reparameterization (Eqn.~\ref{eqn:diffusion})
  \STATE Take gradient descent step on \\
  ~~~ $\nabla_{\bm{\theta}}\left[-\log p_{\bm{\theta}}(\bm{y}|\z{0}) + \| \bm{z}_{\bm{\theta}}(\z{t}, \bm{x}, t)-\z{0}\|_2^2\right]$
  \UNTIL{converged}
\end{algorithmic}
\end{algorithm}

As a result, the updated diffusion loss with an enlarged minimum timestep threshold (thus an increased minimum noise scale) in the final training objective (modified from Eqn. \ref{eqn:difflm_obj}) now becomes:
\begin{equation}
\small
\begin{aligned}
   \mathcal{L}^{\prime}_{\text{diffusion}}(\bm{y}) = \!\!\!\!\!\!\EE_{t \sim \mathcal{U}(t_{\min}, 1), \bm{\epsilon}_t} \Big[\|\bm{z_\theta}(\z{t}, \bm{x}, t) - \bm{z}_0\|_2^2 \Big].
\end{aligned} \nonumber
\end{equation}
We provide pseudocodes regarding how to manipulate noises in training as such in Alg.~\ref{alg:training}.



\subsection{\textsc{CeDi}: Manipulating Noises for Condition-Enhanced Sampling}
\label{subsec: sampling}

\begin{figure}[t]
    \centering
    \includegraphics[width=\columnwidth]{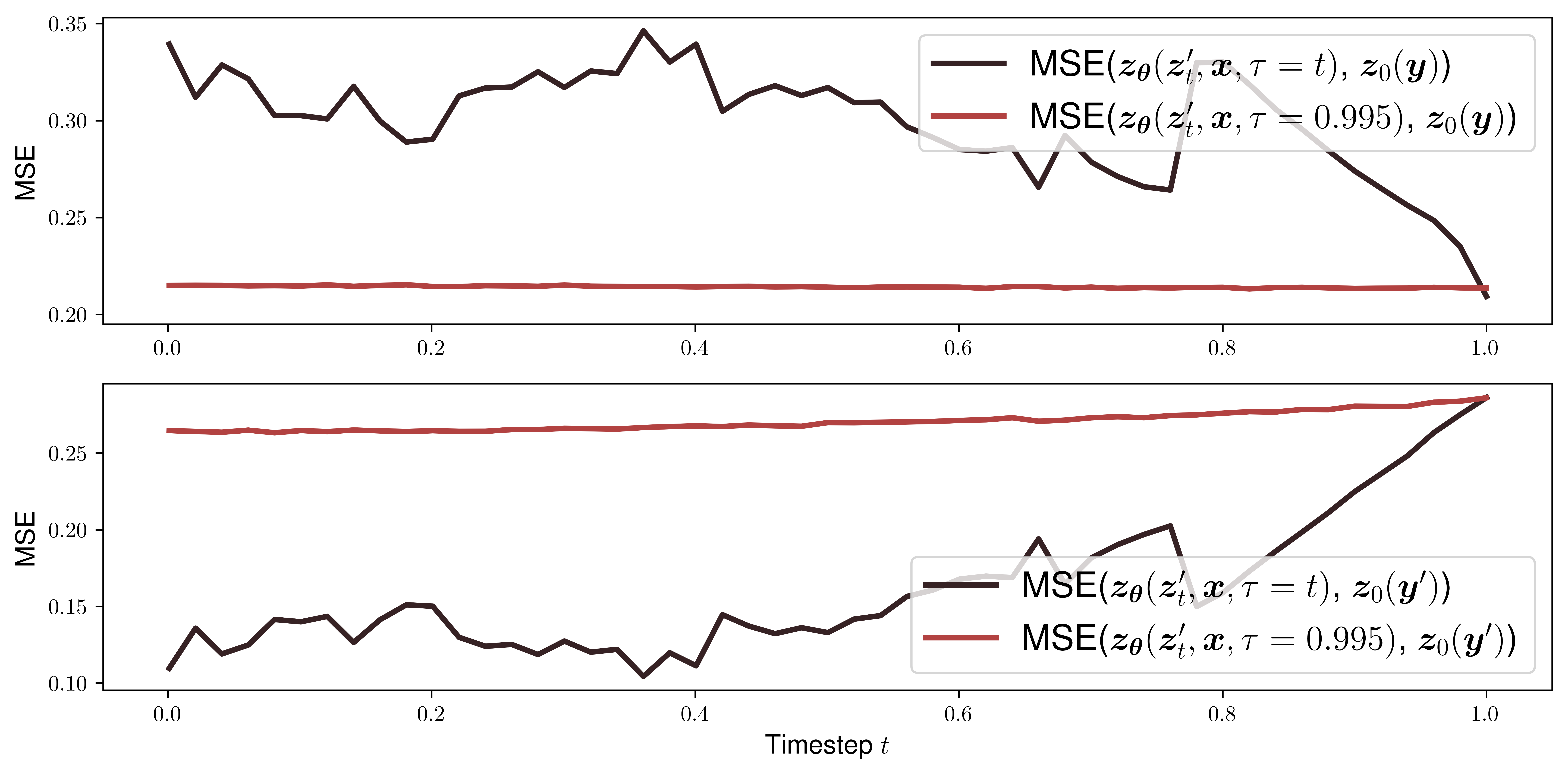}
    \vspace{-7mm}
    \caption{\revise{A synthesis experiment where the model is asked to predict with current timestep $\tau = t$ and an alternative larger timestep $\tau = 0.995$, respectively. 
    We compare the MSE between the model prediction $\bm{z}_{\bm{\theta}}(\bm{z}_t^\prime, \bm{x}, \tau)$ to the embedding of ground-truth $\z{0}(\bm{y})$~(top) and negative sample $\z{0}(\bm{y}^\prime)$ (bottom) respectively, and study to which target the model prediction assimilates, the original $\bm{y}$ or the negative one $\bm{y}^{\prime}$, hence should most likely be governed by the source or the target information. }}
    \label{fig:tendency}
    \vspace{-1mm}
\end{figure}

Based on C3 in \S\ref{sec:problem2}, we suppose the model relies more on the source conditions when the input noise scale is large. 
\revise{This implies that we may make the model generate prediction more faithful to source conditions by feeding timesteps corresponding to large noise scales to the model.
Fig.~\ref{fig:tendency} shows a synthesis experiment similar to Fig.~\ref{fig:prelim}(E), wherein the predictions using a large timestep $0.995$ (namely, a larger noise scale) are closer to the embedding of the original target $\bm{y}$, while more distant to that of the misleading $\bm{y}^\prime$, reiterating that the model relies more on the source condition $\bm{x}$ when being exposed to a larger noise due to manipulation in inference. }


Accordingly, we propose a \uline{\textbf{c}}ondition-\uline{\textbf{e}}nhanced \uline{\textbf{d}}eno\uline{\textbf{i}}ser (\textsc{CeDi}) for sampling.
\textsc{CeDi} always feeds a large $t$ to the model $\bm{z_\theta}$ to encourage the model to make use of the source condition. 
In practice, we largely follow the framework of DDIM solver~\citep{ddim} but pick two sets of timesteps. 
In the first set $\{t_i\}_{i=0}^M$, timesteps decrease uniformly from $t_0=1$ to $t_M\approx0$ as normal. 
As for the other set $\{\tau_i\}_{i=0}^M$, $\tau_i$s decrease uniformly from $\tau_0=1$ to a large time $\tau_M\gg 0$\footnote{Empirically, we find that $\tau_M$ satisfying $\sigma(\tau_M)=0.99$ (\ie, $\tau_M=0.99$ for $\sigma(t)=t$ and $\tau_M=0.9606$ for \citet{diffusionlm}'s \texttt{sqrt} schedule $\sigma(t) = t^{0.25}$) generally works well.}. 
When making predictions, we assign timesteps from the second set to the model.
By replacing corresponding timesteps in the framework of DDIM (Eqn.~\ref{eqn:ddim} and Eqn.~\ref{eqn:predicted-noise}) with $\tau_i$, we generate our predictions by iteratively computing
\begin{equation}\small
\resizebox{\columnwidth}{!}{$
\hat{\bm{z}}_{t_{i}} = \alpha(t_i) \bm{z}_{\bm{\theta}} (\hat{\bm{z}}_{t_{i-1}}, \bm{x}, \textcolor{red}{\tau_{i-1}}) + \sigma(t_{i}) \bm{\epsilon}_{\bm{\theta}}(\hat{\bm{z}}_{t_{i-1}}, \bm{x}, \textcolor{red}{\tau_{i-1}}), \nonumber
$}
\end{equation}
where the predicted noise is also updated as
\begin{equation}
\small
\resizebox{\columnwidth}{!}{$
    \bm{\epsilon}_{\bm{\theta}}(\hat{\bm{z}}_{t_{i-1}}, \bm{x}, \textcolor{red}{\tau_{i-1}})=\frac{\hat{\bm{z}}_{t_{i-1}} - \alpha(\textcolor{red}{\tau_{i-1}})\bm{z}_{\bm{\theta}} (\hat{\bm{z}}_{t_{i-1}},\bm{x}, \textcolor{red}{\tau_{i-1}})}{\sigma(\textcolor{red}{\tau_{i-1}})}. \nonumber
$}
\end{equation}
We also demonstrate how \textsc{CeDi} works in Alg.~\ref{alg:sampling}.

\begin{algorithm}[t]
\small
\caption{Sampling with \method}
\label{alg:sampling}
\textbf{Input} Source condition $\bm{x}$; number of steps $M$; model parameters $\bm{\theta}$. \\
\textbf{Output} Predicted target $\hat{\bm{y}}$.

\begin{algorithmic}[1]
    \STATE Uniformly discretize $[T, 1]$ into $M+1$ steps $\{t_i\}_{i=0}^M$ in descend order ($T\approx0$)
    \STATE Uniformly discretize $[\mathcal{T}, 1]$ into $M+1$ steps $\{\tau_i\}_{i=0}^M$ in descend order ($\mathcal{T}\gg0$)
    \STATE Sample $\hat{\bm{z}}_{t_0}\sim \mathcal{N}(\bm{0}, \bm{I})$
    \FOR{$i=1$ to $M$}
        \STATE $\hat{\bm{z}}_0 \leftarrow \bm{z}_{\bm{\theta}} (\hat{\bm{z}}_{t_{i-1}}, \bm{x}, \textcolor{red}{\tau_{i-1}})$
        \STATE $\hat{\bm{\epsilon}} \leftarrow \frac{\hat{\bm{z}}_{t_{i-1}} - \alpha(\textcolor{red}{\tau_{i-1}})\hat{\bm{z}}_0}{\sigma(\textcolor{red}{\tau_{i-1}})}$
        \STATE $\hat{\bm{z}}_{t_{i}}\leftarrow\alpha(t_i)\hat{\bm{z}}_0 + \sigma(t_{i})\hat{\bm{\epsilon}} $
    \ENDFOR
    \STATE Map $\hat{\bm{z}}_{t_M}$ to $\hat{\bm{y}}$ with the embeddings
\end{algorithmic}
\end{algorithm}

\section{Experiment}
We conduct experiments to verify the effectiveness of the \method and study its characteristics.
\subsection{Experimental Setup}
\label{sec:exp-setup}

\paragraph{Tasks and Datasets.}  
We mainly experiment on machine translation, a well-established benchmark task for conditional sequence learning. We consider IWSLT14 \textsc{De}$\leftrightarrow$\textsc{En}~(160K pairs), WMT14 \textsc{En}$\leftrightarrow$\textsc{De} (4.0M pairs), and WMT14 \textsc{En}$\leftrightarrow$\textsc{Ro} (610K pairs), six translation tasks with variant sizes of training data.
Additionally, we experiment on two of the datasets introduced by DiffuSeq~\citep{gong2022diffuseq}, including Wiki~\citep{data_wiki} for text simplification and QQP\footnote{\url{https://www.kaggle.com/c/quora-question-pairs}} for paraphrasing. 


\vspace{-2mm}
\paragraph{Baselines.} We include three groups of baselines for machine translation:
(1) The autoregressive Transformer~\citep{vaswani2017attention};
(2) The CMLM~\citep{ghazvininejad2019mask}, an iterative-based non-autoregressive model for conditional sequence learning.
(3) Previous diffusion-based sequence generative models, including the vanilla design that simply extends the original DiffusionLM~\citep{diffusionlm} with an additional condition encoder, and the other recently proposed improved methods CDCD \citep[continuous diffusion for categorical data,][]{cdcd}, DiffuSeq~\citep{gong2022diffuseq}, SeqDiffuSeq~\citep{yuan2022seqdiffuseq} and Difformer~\citep{gao2022difformer}. 
For text simplification and paraphrasing, we compare our method with DiffuSeq~\citep{gong2022diffuseq}.

\vspace{-2mm}
\paragraph{Metrics.} We primarily report SacreBLEU\footnote{The signature is 
\texttt{nrefs:1|case:mixed|eff:no|
tok:intl|smooth:exp|version:2.3.1} if the target language is German, and \texttt{nrefs:1|case:mixed|
eff:no|tok:13a|smooth:exp|version:2.3.1} for others.}~\citep{post2018call} for machine translation, following CDCD~\citep{cdcd}.
For text simplification and paraphrasing, we follow DiffuSeq to employ sentence-level BLEU under the tokenizer of \texttt{BERT-BASE-UNCASED}. 

\begin{table*}[t]
\small
\vspace{-3mm}
\caption{Comparison in \textbf{SacreBLEU} on machine translation tasks. ``LB'': the size of the length beam search.
``MBR'': the number of candidates for each length beam to apply Minimum Bayes-Risk decoding. 
``\textbf{KD}'': results are obtained with knowledge distillation~\citep[KD,][]{kim2016sequence,zhou2020understanding}.
Provided that KD is common and effective practice in non-autoregressive (NAR) machine translation, though not the focus of this study, we also provide further experiments with KD for reference. 
The best NAR results with and without KD are in \textbf{bold} and the second best ones are \underline{underlined}. 
\revise{We report 95\% confidential interval for our method computed with \texttt{compare-mt}~\citep{compare-mt}}.\\
$\dag$: how CMLM originally selects candidates with different lengths differs from the MBR decoding we used for diffusion models, and we thus include its results with MBR decoding for fair comparisons.
$\ddag$: the results are quoted from \citet{cdcd}.
$\dddag$: the results are quoted from \citet{gao2022difformer} while the results of the rest datasets are missing in the original paper, for which we obtain through their opensource code. Note that the results of DiffuSeq and SeqDiffuSeq are presented in tokenized BLEU as reported in \citet{gao2022difformer}, and we encourage readers to check the original papers for more details.   }
\label{tab:mt-result}
\centering
\setlength{\tabcolsep}{6pt}
\resizebox{\linewidth}{!}{
\begin{tabular}{lcccccc}
\toprule
\multirow{2}{*}{\textbf{Methods}} & \multicolumn{2}{c}{\textbf{IWSLT14}}& \multicolumn{2}{c}{\textbf{WMT14}}& \multicolumn{2}{c}{\textbf{WMT16}} \\

\cmidrule[0.4pt](lr){2-7} 

& \multicolumn{1}{l}{{\textsc{De}$\rightarrow$\textsc{En}}} 
& \multicolumn{1}{l}{{\textsc{En}$\rightarrow$\textsc{De}}} 
& \multicolumn{1}{l}{{\textsc{De}$\rightarrow$\textsc{En}}} 
& \multicolumn{1}{l}{{\textsc{En}$\rightarrow$\textsc{De}}} 
& \multicolumn{1}{l}{{\textsc{Ro}$\rightarrow$\textsc{En}}} 
& \multicolumn{1}{l}{{\textsc{En}$\rightarrow$\textsc{Ro}}} \\ 
\midrule

Transformer~\smallcitep{vaswani2017attention}~~(AR, beam $=5$) & 33.61    & 28.30    & 30.55    & 26.85 &  33.08    & 32.86 \\ 
\cmidrule[0.4pt](lr){1-7}
CMLM~\smallcitep{ghazvininejad2019mask}~~(NAR, LB $=5$)   & 29.41    & 24.33    & 28.71    & 23.22 &  31.13    & \textbf{31.26}  \\ 
CMLM~\smallcitep{ghazvininejad2019mask}~~(NAR, LB $=5$, MBR=1$^\dag$)   & 29.32    & 24.34    & 28.43    & 23.09 &  31.07    & 30.92  \\ \midrule
DiffusionLM~\smallcitep{diffusionlm}~~(LB $=5$, MBR $=1$)    & 26.61    & 20.29    & 17.31    & 15.33 &  28.61    & 27.01   \\
DiffusionLM~\smallcitep{diffusionlm}~~(LB $=5$, MBR $=10$)   & 29.11    & 22.91    & 19.69    & 17.41 &  30.17    & 29.39    \\
CDCD~\smallcitep{cdcd}~~~(MBR $=10$)   & - & - & 25.40$^\ddag$    & 19.70$^\ddag$ & -  &  -    \\
CDCD~\smallcitep{cdcd}~~~(MBR $=100$)  & - & - & 26.00$^\ddag$    & 20.00$^\ddag$ &   -   &  -    \\ 
Difformer~\smallcitep{gao2022difformer}~~(LB $\times$ MBR $=20$) & \revise{28.01} & \revise{23.31} & \revise{25.30} & 23.80$^{\dddag}$ &   \revise{29.37}   &  \revise{29.20}    \\ 

\cmidrule[0.4pt](lr){1-7}

\chl \method~~(LB $=5$, MBR $=1$)    & \chl 31.29\revise{$_{0.67}$} & \chl 25.55\revise{$_{0.65}$} & \chl 28.83\revise{$_{0.92}$}             & \chl 24.25\revise{$_{0.86}$}             &\chl 31.14\revise{$_{1.13}$}                &\chl  30.93\revise{$_{1.12}$}  \\

\chl \method~~(LB $=5$, MBR $=10$)   &\chl \textbf{31.61}\revise{$_{0.67}$}    &\chl \underline{25.70}\revise{$_{0.62}$}    &\chl \textbf{29.05}\revise{$_{0.92}$}    &\chl \underline{24.26}\revise{$_{0.84}$} &\chl \underline{31.22}\revise{$_{1.15}$}    &  \chl \underline{31.08}\revise{$_{1.12}$}   \\

\chl \method~~(LB $=10$, MBR $=5$)   &\chl \underline{31.44}\revise{$_{0.68}$}    &\chl  \textbf{26.14}\revise{$_{0.65}$}   &\chl \underline{29.01}\revise{$_{0.88}$} &\chl  \textbf{24.62}\revise{$_{0.88}$}   &\chl \textbf{31.24}\revise{$_{1.12}$}        &\chl 31.03\revise{$_{1.13}$} \\

\midrule

DiffuSeq~\smallcitep{gong2022diffuseq}~~~~~~~(\textbf{KD}, LB$\times$MBR $=10$) & - & - & -    & 15.37$^{\dddag}$ &   -   &  25.45$^{\dddag}$    \\ 
SeqDiffuSeq$^{\dddag}$~\smallcitep{yuan2022seqdiffuseq}~~(\textbf{KD}, LB$\times$MBR $=10$) & - & - & -    & 17.14$^\ddag$ &   -   &  26.17$^{\dddag}$    \\ 
\cmidrule[0.4pt](lr){1-7}
\chl \method~~(\textbf{KD}, LB $=10$, MBR $=5$)   &\chl -   &\chl  -   &\chl \bf {30.30\revise{$_{0.94}$}}    &\chl \bf  {25.88\revise{$_{0.95}$}}    &\chl \bf  33.13{\revise{$_{1.20}$}}   &\chl \bf  {32.84\revise{$_{1.16}$}}\\

\bottomrule
\end{tabular}
}%
\vspace{-3mm}
\end{table*}

\vspace{-2mm}
\paragraph{Implementation.}
All our implementations are based on \texttt{Transformer-BASE}~\citep{vaswani2017attention} for all datasets except IWSLT14. 
For IWSLT14, we use a smaller architecture that has 4 attention heads and 1024-dimensional feedforward layers. 
The embedding dimension for the diffusion model is 16 on IWSLT14 and 64 on the others. 
In the implementation of our method, we follow recent advances and apply self-conditioning techniques~\citep{cdcd,chen2022analog,strudel2022self}. 
Besides, following previous practice in non-autoregressive machine translation, we train our model both with and without knowledge distillation\footnote{\revise{Non-autoregressive sequence learning models typically struggle with learning multimodal distributions~\citep{gu2018non}. For this reason, a common technique to improve their performance is to apply knowledge distillation, which simplifies the target distribution by replacing target samples with predictions from an autoregressive teacher model.}}~\citep[KD,][]{kim2016sequence,zhou2020understanding}.

During inference, for machine translation, we apply beam search in the autoregressive Transformer with beam size~5.
Correspondingly, we use length beam 5 in the non-autoregressive models, except for CDCD and DiffuSeq since they vary the target lengths by predicting paddings instead of length predictions.
For text simplification and paraphrasing, we report results with various length beams as length prediction on these tasks is more challenging and less studied.
For all the diffusion-based methods, we follow previous work~\citep{diffusionlm,gong2022diffuseq,cdcd} and apply Minimum Bayes-Risk (MBR) decoding~\citep{kumar2004minimum}. 
For both DiffusionLM and our model, we perform sampling with 20 steps.

We implement DiffusionLM and \method upon \code{fairseq}~\citep{ott2019fairseq}, and also train Transformer and CMLM baselines using \code{fairseq}.
The training batch size is 128K for WMT14/WMT16, and 32K for the others.
For more details, please refer to \S\ref{sec: implementation details}.


\subsection{Main Results}
The results of machine translation and the other two tasks are in Tab.~\ref{tab:mt-result} and Tab.~\ref{tab:diffuseq-data}, respectively. 

\paragraph{Overall performance.} 
Our \method demonstrates effectiveness on all selected conditional sequence learning tasks,
which we summarize into the following three aspects:
\begin{compactitem}
    \item \method achieves state-of-the-art results among diffusion-based models on one of the representative conditional sequence generation tasks, \ie, machine translation, where \method outperforms the vanilla design of DiffusionLM, as well as the previous strongest approaches such as CDCD~\citep{cdcd} and Difformer~\citep{gao2022difformer} by a large margin (Tab.~\ref{tab:mt-result}). 
    For DiffuSeq and SeqDiffuSeq, although their reported tokenized BLEUs are not strictly comparable to our SacreBLEU results due to the difference in tokenizers, our performance is far above them by over 4 BLEU score and even more if we involve knowledge distillation, which supports our superiority over them.
    
    \item \method demonstrates strong competitiveness in conditional sequence learning.
    It even surpasses CMLM~\citep{ghazvininejad2019mask} on almost all the experimented machine translation datasets~(Tab.~\ref{tab:mt-result}). 
    Provided that CMLM is one of the leading approaches among NAR sequence learners, the performance \method achieves can be considered quite competitive.
    
    \item \method is generic to various conditional sequence learning tasks. Results on Tab.~\ref{tab:diffuseq-data} shows that \method also works well in tasks other than machine translation, surpassing previously proposed DiffuSeq.
\end{compactitem}



In addition to the overall performance, \method also demonstrates several nice properties. 
We elaborate on them as follows:

\paragraph{Scalability.} 
As shown in Tab.~\ref{tab:mt-result}, DiffusionLM seems more challenging to accommodate larger datasets like WMT14 than smaller ones (\eg, IWSLT14). 
This verifies the curse of scalability problem discussed in \S\ref{sec:preliminary}.
In contrast, \method surpasses CMLM on almost all large- and small-scale scenarios, which indicates that \method indeed greatly improves the scalability of diffusion-based sequence learners. 
This advantage of \method could help facilitate further research of large-scale real-world applications of diffused sequence generative models.

\paragraph{Sampling efficiency.}  
Given the sizes of length beam (LB) and MBR decoding shown in Tab.~\ref{tab:mt-result}, \method surpasses or closely approaches CMLM even when MBR=$1$, while the vanilla DiffusionLM heavily relies on a large number of candidates for MBR decoding.
Besides, \method necessitates much fewer NFEs to achieve strong performance, \eg, only 20 steps, resulting in only 1\% to 10\% computational costs and latency compared to previous works~\citep{gong2022diffuseq,diffusionlm,cdcd}.
This manifests that \method is more accurate yet efficient compared to previous diffusion-based sequence learning models.

\begin{table}[t]
\vspace{-3mm}
\caption{\textbf{Sentence-level BLEU} of our method and DiffuSeq on Wiki (text simplification) and QQP (paraphrasing). 
``NFE'': number of function evaluations, measuring the total number of forward passes to the model for each prediction.
The results of DiffuSeq are quoted from~\citet{gong2022diffuseq}.
}
\label{tab:diffuseq-data}
\resizebox{\columnwidth}{!}{
\small
\centering
\tabcolsep 4.25pt
\begin{tabular}{lrrrrcc}
    \toprule
    \textbf{Methods} & \textbf{Steps}& \textbf{LB} & \textbf{MBR} &\textbf{NFE}& \textbf{Wiki} & \textbf{QQP} \\ 
    \midrule
    DiffuSeq & 2000    & -~ & 10    & 20000 & 36.22  & 24.13    \\
    \midrule
    \method & 20  & 10 & 1 & 200   & 35.36\revise{$_{1.63}$}  & \textbf{26.07}\revise{$_{1.24}$}  \\
    \method & 20  & 20 & 1 & 400   & \textbf{36.94}\revise{$_{1.95}$}   & 25.42\revise{$_{1.46}$}    \\
    \method & 20  & 20 & 5 & 2000  & 36.88\revise{$_{1.77}$}  & 25.57\revise{$_{1.29}$}    \\ 
    \bottomrule
\end{tabular}
}%
\vspace{-3mm}
\end{table}

\begin{table*}[t]
\vspace{-2mm}
\caption{Ablation Study on WMT14 \textsc{En}$\rightarrow$\textsc{De}. All the results are in SacreBLEU scores with $LB=5$.}
\label{tab:ablation}
\centering
\small
\tabcolsep 3pt
\resizebox{0.93\linewidth}{!}{
\begin{tabular}{lcccc} 
\toprule
\bf Training Settings & DDIM (MBR $=1$)   & \textsc{CeDi} (MBR $=1$) & DDIM (MBR $=10$)    & \textsc{CeDi} (MBR $=10$)    \\ 
\midrule
Ours [final]                                        & 19.23 & 24.25   & 22.12 & 24.26  \\
w/o self-conditioning                               & 20.37 & 23.03   & 22.58 & 23.14  \\
w/o noise scale clipping                          & 7.95  & 21.16   & 11.51 & 21.40  \\
w/o self-conditioning, w/o noise scale clipping   & 11.30 & 20.86   & 14.84 & 21.47  \\ 
\midrule
w/ sqrt noise schedule                              & 20.11 & 24.13   & 22.83 & 24.07  \\
w/ sqrt noise schedule, w/o noise scale clipping  & 16.68 & 23.22   & 20.46 & 23.40  \\
\bottomrule
\end{tabular}
}
\vspace{-2mm}
\end{table*}

\vspace{-2mm}
\subsection{Effect of Noise Scale Clipping for Training}
\vspace{-1mm}
\paragraph{Ablation study on noise clipping.}
We compare models trained with different settings in Tab.~\ref{tab:ablation} to study the effect of our training strategy. 
We find that the proposed noise scale clipping consistently helps improve the model performance. 
Replacing the noise schedule in \method (\ie, $\sigma(t) = t)$ with the \texttt{sqrt} schedule proposed by~\citet{diffusionlm} has negligible influence on the final performance. 
This is expected since we only set the noise schedule as $\sigma(t)=t$ for convenience. 
The difference is that the improvement by noise clipping is relatively smaller when using the \texttt{sqrt} schedule. 
This is because the noise scale of the sqrt schedule increases quickly in small timesteps (Fig.~\ref{fig:noise-clipping}B). When $t=0.2$, $\sigma_{sqrt}(t)\approx 0.67$, which is close to our initial clipping threshold. 
This suggests the success of the sqrt schedule may also be partly explained as involving more large-scale noises.

\begin{figure}[t]
    \vspace{-2mm}
    ~~~~\includegraphics[width=0.86\linewidth]{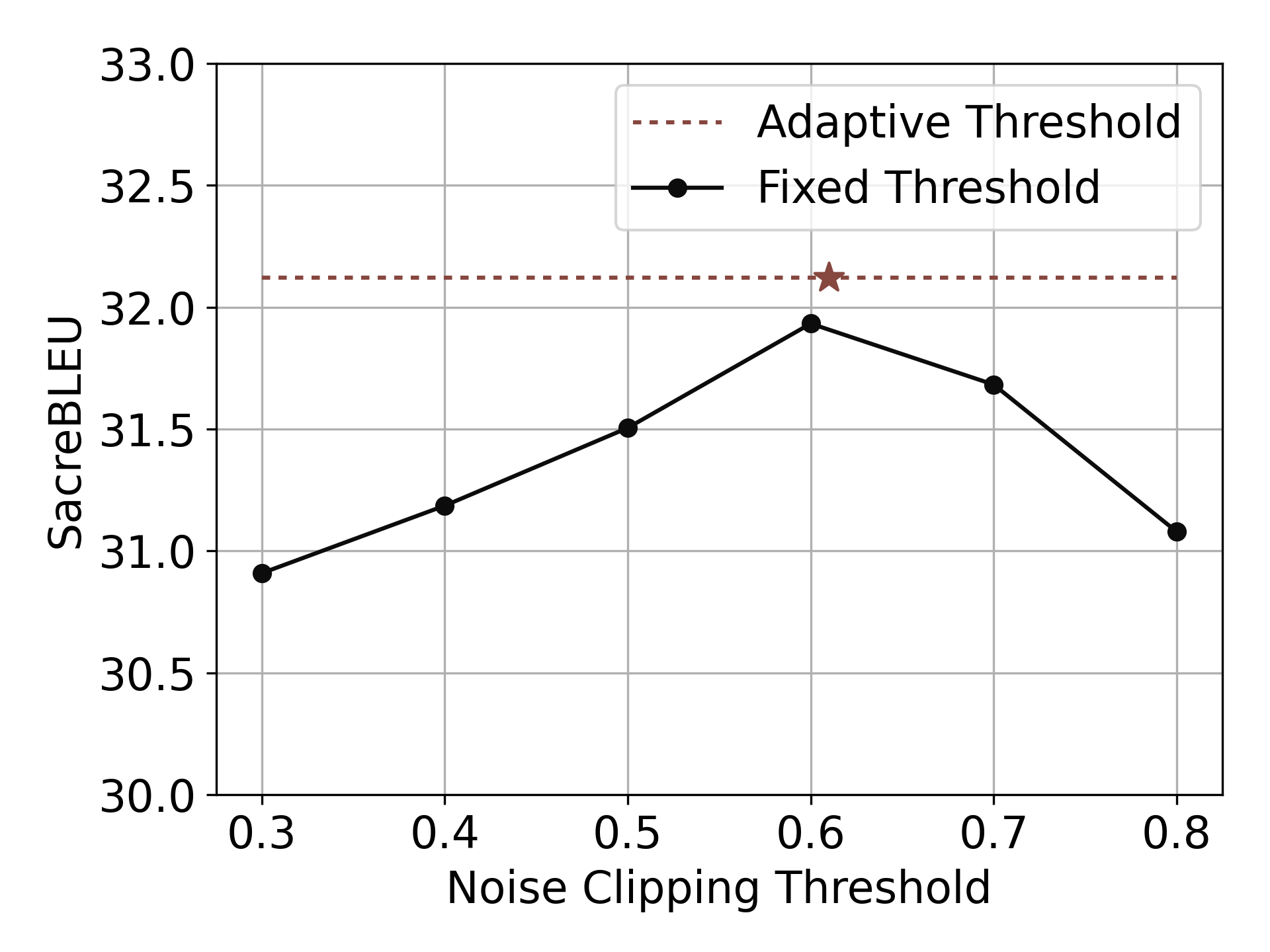}
    \vspace{-4mm}
    \caption{\revise{SacreBLEU on IWSLT14 \textsc{De}$\rightarrow$\textsc{En} with our models trained with adaptive \vs different fixed noise clipping thresholds. We sample results with MBR=5 and oracle length. The star marker~($\star$) stands for the clipping threshold of our final checkpoint trained with adaptive clipping threshold.}}
    \vspace{-3mm}
    \label{fig:thres}
\end{figure}

\vspace{-2mm}
\paragraph{On the effect of different noise scale clipping thresholds.}
\revise{To understand how different noise clipping thresholds affect the model performance, we compare our model trained with adaptive noise threshold and different fixed noise thresholds in Fig.~\ref{fig:thres}.
Results show that the performance degrades if the clipping threshold is either too small or too large. 
Our proposed strategy adaptively finds the balance well.
It finds the ideal noise clipping threshold and achieves strong performance.
}

\subsection{Effect of Condition Enhancement for Sampling} 
\begin{figure}[t]
    \centering
    \includegraphics[width=0.95\linewidth]{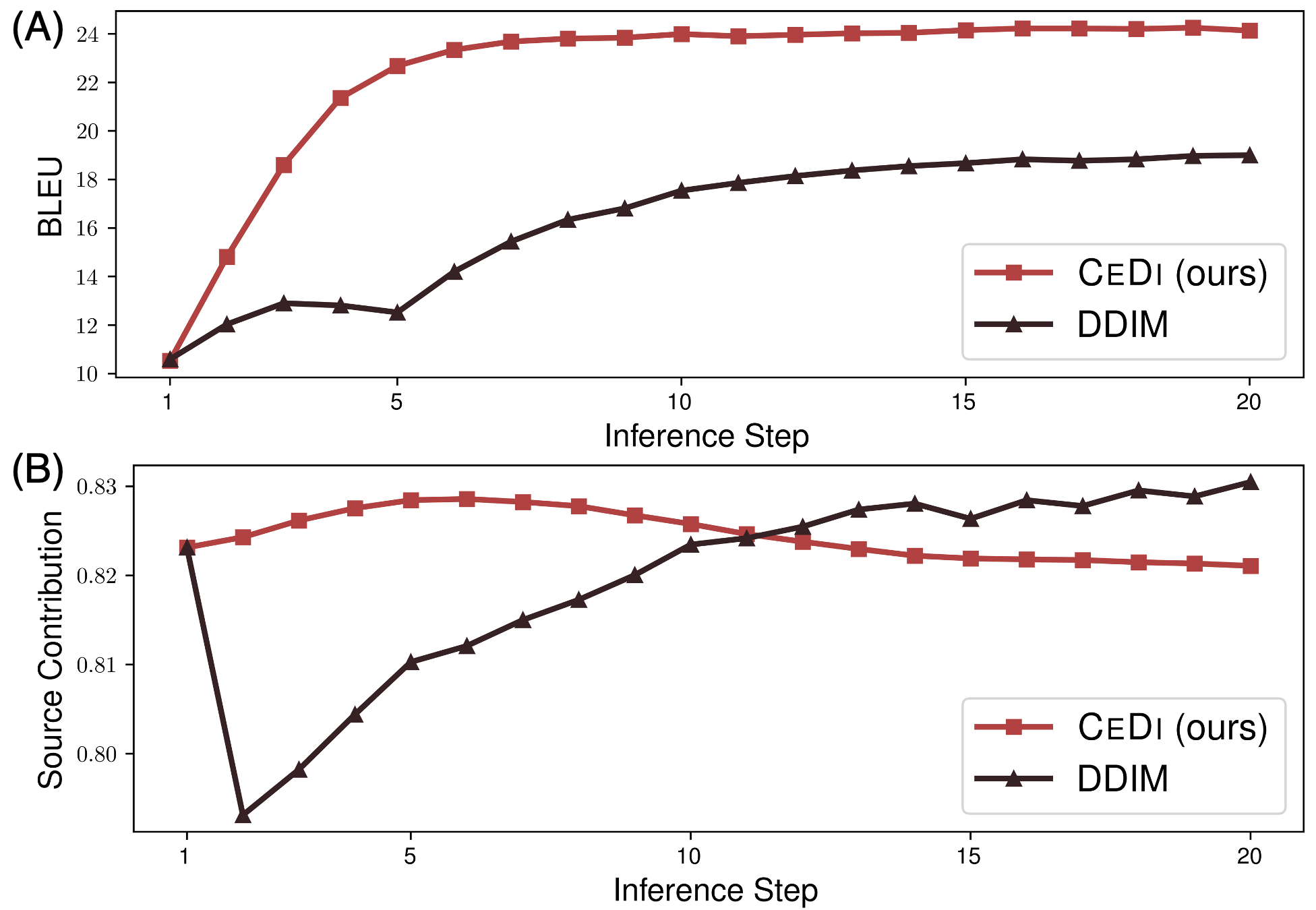}
    \vspace{-3mm}
    \caption{The difference between \textsc{CeDi} and DDIM solver over steps. (A) The prediction accuracy at each step, measured with SacreBLEU. 
    (B) The proportion of source contribution to the prediction \revise{in Layer-wise Relevant Propagation (LRP)} at each step. }
    \label{fig:lrp}
    \vspace{-5mm}
\end{figure}

We compare the performance between different denoisers, \ie, DDIM and the proposed \textsc{CeDi}, in Tab.~\ref{tab:ablation}.
In a nutshell, \textsc{CeDi} impressively outperforms DDIM, especially for small MBR candidate sizes.
We also notice that DDIM performs particularly unsatisfactorily when the model is trained without noise scale clipping. 
However, for these models, \textsc{CeDi} can still produce a relatively good performance (over 20.50 for MBR=1) that even surpasses well-designed CDCD (20.00 for MBR=100). 
What's more, we highlight two critical characteristics of \textsc{CeDi} as follows:

\vspace{-2mm}
\paragraph{\textsc{CeDi} indeed better leverage source conditions for inference.} Recall that we propose the \textsc{CeDi} with the purpose of encouraging the model to make better use of source conditions for prediction (\S\ref{subsec: sampling}). 
To investigate whether the denoiser achieves this, we apply Layer-wise Relevant Propagation~\citep[LRP,][]{bach2015pixel,voita2021analyzing} to measure the relative contribution of the source condition to the model prediction. 
As shown in Fig.~\ref{fig:lrp}(B), we compare the source contribution of our \textsc{CeDi} and the DDIM solver along the sampling iterations.
\textsc{CeDi} maintains a high source contribution, while the source contribution of \textsc{CeDi} is unsatisfactory in the first few steps, which demonstrates that sampling with our \textsc{CeDi} does leverage the source condition more. 
Correspondingly, as shown in Fig.~\ref{fig:lrp}(A), the prediction accuracy of \textsc{CeDi} increases steadily, while the performance of DDIM fails to improve at the beginning of the iterations, suggesting correlations between higher source contribution and higher performance improvement.
Among all iterations, the first few steps establish the foundation for the overall performance. 
Although DDIM improves its performance in later iterations, it still falls behind our \textsc{CeDi}. 
This suggests the effectiveness of increasing the source contribution, especially at the beginning of the sampling process.

To further show the strength of \textsc{CeDi} in capturing source conditions, we explore more complex conditional sequence learning scenarios. 
We simulate this under two multilingual translation settings, \ie \texttt{many-to-one} and \texttt{one-to-many} translation. 
In the \texttt{many-to-one} scenario, a unified model needs to translate source sentences in multiple different source languages to English counterparts, requiring the model to handle complicated source conditions.
On the other hand, the \texttt{one-to-many} setting simulates a multi-conditional scenario, requiring the model to recognize the target language as another crucial condition to capture the target distribution.
To this end, we construct a dataset by combining four language pairs of
IWSLT14 translation benchmark, \ie, \textsc{En}$\leftrightarrow$\textsc{De}, \textsc{En}$\leftrightarrow$\textsc{Ro}, \textsc{En}$\leftrightarrow$\textsc{Nl}, and \textsc{En}$\leftrightarrow$\textsc{Pt-br}. 
In the \texttt{one-to-many} translation, we append language tokens to the source sequences to incorporate the target language as a condition. 
We also include a baseline in which the models are trained separately for each language pair for comparison.

\begin{table}[t]
\vspace{-2mm}
\caption{Results of multilingual machine translation (\{\textsc{De},\textsc{Ro},\textsc{Pt-br},\textsc{Nl}\}$\leftrightarrow$\textsc{En}).
``\textsc{Bilingual}'': integrated results of separate models of every language pair.
``\textsc{Multiling.}'': results from a unified multilingual model.
We employ \texttt{langdetect}~\citep{langdetect} to infer the language of generated sequences for computing the language accuracy.}
\label{tab:multilingual}
\small
\centering
\setlength{\tabcolsep}{2pt}
\resizebox{\columnwidth}{!}{
\begin{tabular}{@{}llcc@{}}
\toprule
\multirow{2}{*}{\textbf{Settings}}         
& \multirow{2}{*}{\textbf{Methods}}         
& \multicolumn{2}{c}{\textbf{SacreBLEU (Lang Acc \%)}} \\
& & \textsc{Bilingual}   & \textsc{Multiling.}       \\ 
\midrule
\multirow{5}{*}{\shortstack{\texttt{many-to-one}\\\\\{\textsc{De},\textsc{Ro},\textsc{Pt-br},\textsc{Nl}\}\\$\to$\textsc{En}}} 
& CMLM            & 33.85        &  35.23~~~~~~~~~~~~~ \\
& \method (DDIM, MBR=1)  & 32.40 &  33.43~~~~~~~~~~~~~    \\
& \method (DDIM, MBR=10) & 33.73 &  35.48~~~~~~~~~~~~~   \\ 
&\chl \method (MBR=1)  &\chl 34.57       &\chl  35.26~~~~~~~~~~~~~    \\
&\chl \method (MBR=10) &\chl 34.74       &\chl  35.66~~~~~~~~~~~~~  \\ 

\midrule
\multirow{5}{*}{\shortstack{\texttt{one-to-many}\\\\\textsc{En}$\to$\\\{\textsc{De},\textsc{Ro},\textsc{Pt-br},\textsc{Nl}\}}} 
& CMLM            & 28.10 & 30.55 (94.73) \\
& \method (DDIM, MBR=1) & 27.44 & 17.95 (89.77)         \\ 
& \method (DDIM, MBR=10) & 28.54 & 18.57 (89.53)         \\ 
&\chl \method (MBR=1)  &\chl 28.72 &\chl 30.52 (95.31) \\
&\chl \method (MBR=10) &\chl 28.81 &\chl 30.67 (95.42) \\ 
\bottomrule
\end{tabular}
}
\vspace{-3mm}
\end{table}

\vspace{-3mm}
\paragraph{\textsc{CeDi} can handle sequence generation from complex and multiple conditions.}
\begin{table*}[t]
\vspace{-2mm}
\caption{\revise{A quanlitative example for one-to-many translation. The source contains both the English sentence to be translated and the target language. We compare generation results of CMLM, \method but sampling with DDIM instead of \textsc{CeDi}, and the complete \method.}  }
\label{tab:qualitative}
\centering
\small
\tabcolsep 4pt
\resizebox{\linewidth}{!}{
\begin{tabular}{ll}
\toprule
\textbf{Source}             & (target language: ro) something as dramatic as our identity has now become a matter of choice, as this slide is meant to indicate. \\ \midrule
\textbf{Reference}          & ceva atât de important ca identitatea noastră a devenit acum o problemă de alegere, și această tranziție are rolul de a arăta ast  \\ \midrule
\textbf{CMLM}               & ceva la fel de dramatic ca identitatea noastră a devenit o problemă de alegere, cum acest slide este făcut să arate.               \\
\textbf{\method (w/ DDIM)} & ceva de de de de de de de de de a de de de de de de de de de de de de de de de de.                                                 \\
\textbf{\method (w/ \textsc{CeDi}})           & ceva atât de dramatic ca identitatea noastră, a devenit acum o problemă de alegere, aşa cum se înseamnă să indice acest imagine    \\ 
\bottomrule
\end{tabular}
}
\vspace{-3.5mm}
\end{table*}

As shown in Tab.~\ref{tab:multilingual}, \method works well in multilingual settings, showing its strong capability in modeling conditions, \ie a complex multimodal condition (source sentences of four languages in \texttt{many-to-one}), and multiple conditions (source sentence in English as well as the identity of target languages).
Particularly, \textsc{CeDi} shows huge advantages over DDIM in the multilingual setting of \texttt{one-to-many} translation. 
In this case, the language accuracy of DDIM is much lower than that of \textsc{CeDi}, suggesting DDIM has trouble capturing the given condition, namely the language identity in this \texttt{one-to-many} scenario.
In contrast, \method augmented with \textsc{CeDi} yields satisfactory predictions with high language accuracy, exhibiting superiority in working with multiple conditions.
\revise{This is also consistent with our findings from the qualitative examples as shown in Tab.~\ref{tab:qualitative}, where DDIM may fail to capture the language condition and generate non-sense articles shared across languages, while the full \method produces fluent and satisfactory results.}

\vspace{-2mm}
\section{Related Work}
\vspace{-8mm}
\label{sec:related}

\paragraph{Non-autoregressive Sequence Generative Models.}
Non-autoregressive sequence learning (NAR) was first proposed by~\citet{gu2018non} as an alternative to its autoregressive counterpart. 
It generates target tokens in parallel, either fully NAR~\citep{gu2018non} or up to a mild number of iterations~\citep{ghazvininejad2019mask}, liberating sequence modeling from the constraint of a predefined order~\citep{qian2022diff}.
With recent efforts, NAR shows great potential in the applications of various domains, including language~\citep{qian2021volctrans,pmlr-v139-qi21a,huang2022directed,qian2022diff}, speech~\citep{kim2021conditional}, proteins~\citep{zheng2023structure,wang2024dplm}, and molecules~\citep{hoogeboom2022equivariant}.
Different from more commonly-used autoregressive (AR) models~\citep{sutskever2014seq2seq}, NAR models assume conditional independence among the output tokens.
Such an assumption risks ignoring the target dependencies~\citep{ren2020study,huang2022learning} and leads to the multi-modality problem~\citep{gu2018non}.
As a result, the vanilla fully NAR model has inferior generation quality.
Some of the later improvements alleviate the strong assumption by reformulating NAR formulation under iterative refinement~\citep{lee2018deterministic,gu2019levenshtein,ghazvininejad2019mask,ghazvininejad2020semi,huang2022non,huang2022improving,zheng2023deep,ye2023diffusion}, which iteratively takes as input the previously generated sequence, which serves as an intermediate random variable, to produce the tokens of its refined or denoised version in parallel until convergence or the budget of maximum iterations run out.
Some recent advances herein follow the idea of discrete diffusion \citep{diffusion2015,austin2021structured} and formalize iterative refinement as Markov processes~\citep{savinov2021step,he2022diffusionbert,reid2022diffuser}. 
Although both are named after diffusion models, these works operate on discrete state space, whereas our focus, continuous diffusion models accommodate the continuous (embedding) space of discrete tokens.

\paragraph{Diffusion Models for Sequence Learning.}
Continuous diffusion models~\citep{diffusion2015,ddpm,song2020sde} gained first success in generating high-quality images.
Recently, DiffusionLM~\citep{diffusionlm} successfully adapted them to sequence learning and proposed the DiffusionLM, the first diffusion-based sequence generative model with a special focus on controllable text generation.
Later improvements to the diffusion-based sequence generative models are mainly categorized threefold.
The first line includes novel components for diffusion modeling, such as the partial diffusion process proposed by DiffuSeq~\citep{gong2022diffuseq}, self-conditioning techniques introduced by \citet{strudel2022self}, and the adaptive noise schedule of \citet{yuan2022seqdiffuseq}.
The second line applies diffusion models to the latent space of specific pretrained language models~\citep{lovelace2022latent}. 
And the third tries to incorporate conventional practice in discrete token prediction. For instance, CDCD~\citep{cdcd}, Difformer~\citep{gao2022difformer} and SSD~\citep{han2022ssd} incorporate the cross-entropy objectives in training.
For the application of diffusion-based models for sequence learning, previous work found their advantages in controllable generation~\citep{yu2022latent,liu2022composable,diffusionlm}, and generating diverse sequences~\citep{gong2022diffuseq}. 
GENIE~\citep{lin2022genie} demonstrates that diffusion-based sequence generative models can benefit from large-scale self-supervised pretraining.
While almost all these works mainly focus on the training phrase of diffusion-based sequence generative models, our study emphasizes both training and inference.

\section{Conclusion}
In this paper, we shed light on the crucial role of noise schedules in diffusion models for conditional sequence learning by systematic empirical study.
Motivated by our findings, we propose \method to determine the best-suited noise scales for both training and inference.
As a result, \method makes training more effective and also enables the model to better utilize source conditions for prediction, thereby leading to considerable performance improvements.
We expect that our study can help facilitate further research on diffusion models to empower various applications in NLP.

\section*{Acknowledgement}
We would like to thank the anonymous reviewers and editors for their invaluable feedback.

\bibliography{tacl2021}
\bibliographystyle{acl_natbib}
\appendix
\onecolumn
\section{More Results on Machine Translation}
\label{sec:more-mt}
To provide references for further study and comparisons, we report more results on machine translation.

\paragraph{Knowledge distillation (KD).} 
A common practice to improve the performance of non-autoregressive machine translation is knowledge distillation~\citep{kim2016sequence,zhou2020understanding}. 
We report the performance of our method trained on distilled data of WMT14 and WMT16 on Tab.~\ref{tab:mt-sacre-kd}. 
The result shows that the performance gap between our method and the autoregressive transformer is small when knowledge distillation is used. 
This suggests that our method achieves performance that satisfies the needs of applications.

\begin{table*}[htbp]
\caption{Model performances on machine translation with knowledge distillation. The results of transformer are from raw data, while \method is trained on distilled data. 
The performances are measured with \textbf{SacreBLEU}.
}
\label{tab:mt-sacre-kd}
\small
\centering
\begin{tabular}{lcccc}
\toprule
    & \multicolumn{2}{c}{\textbf{WMT14}}  
    & \multicolumn{2}{c}{\textbf{WMT16}} \\
\textbf{Methods}   & \multicolumn{1}{l}{\textbf{\textsc{De}$\rightarrow$\textsc{En}}} & \multicolumn{1}{l}{\textbf{\textsc{En}$\rightarrow$\textsc{De}}}  & \multicolumn{1}{l}{\textbf{\textsc{Ro}$\rightarrow$\textsc{En}}} & \multicolumn{1}{l}{\textbf{\textsc{En}$\rightarrow$\textsc{Ro}}} \\ \midrule
Transformer        & 30.55    &  26.85    & 33.08    & 32.86  \\
\method (LB=5, MBR=1)    & 30.13    &  25.70    & 32.96    &  32.58  \\
\method (LB=5, MBR=10)   & 30.12    &  25.90    & 33.04    &  32.57   \\
\method (LB=10, MBR=5)   & 30.30    &  25.88    &  33.13   &  32.84    \\
\bottomrule
\end{tabular}
\end{table*}

\paragraph{Evaluation with tokenized BLEU.} 
Some of the previous studies in machine translation reported tokenized BLEU, despite inconsistent tokenizers (other than the standard Moses tokenizer) they might use.
To help conveniently compare \method to them, we also report the performance of \method with tokenized BLEU\footnote{\url{https://github.com/alvations/sacremoses}} in Tab.~\ref{tab:mt-tokenized}.

\begin{table}[htbp]
\caption{Tokenized BLEU of our method on machine translation datasets. 
We use the \code{moses} tokenizer for all the texts. 
``LB'': the size of length beam.
``MBR'': the number of candidates for each length beam to apply Minimum Bayes-Risk decoding. 
+KD means the results are obtained with knowledge distillation.}
\label{tab:mt-tokenized}
\small
\centering
\resizebox{\linewidth}{!}{
\begin{tabular}{lcccccc}
\toprule
    & \multicolumn{2}{c}{\textbf{IWSLT14}}   & \multicolumn{2}{c}{\textbf{WMT14}}  
    & \multicolumn{2}{c}{\textbf{WMT16}} \\
\textbf{Methods}   & \multicolumn{1}{l}{\textbf{\textsc{De}$\rightarrow$\textsc{En}}} & \multicolumn{1}{l}{\textbf{\textsc{En}$\rightarrow$\textsc{De}}} & \multicolumn{1}{l}{\textbf{\textsc{De}$\rightarrow$\textsc{En}}} & \multicolumn{1}{l}{\textbf{\textsc{En}$\rightarrow$\textsc{De}}}  & \multicolumn{1}{l}{\textbf{\textsc{Ro}$\rightarrow$\textsc{En}}} & \multicolumn{1}{l}{\textbf{\textsc{En}$\rightarrow$\textsc{Ro}}} \\ \midrule
\method (LB=5, MBR=1)    & 32.23    & 25.54    & 29.35    &  24.43    & 31.21    &  31.18  \\
\method (LB=5, MBR=10)   & 32.48    & 25.68    & 29.53    & 24.45 & 31.39    &  31.29   \\
\method (LB=10, MBR=5)   & 32.25 &   25.99   & 29.40    & 24.48 & 31.50    &  31.27   \\
\method + KD (LB=5, MBR=1)   & -    & -    & 30.64    & 26.08 & 33.21    &  32.57   \\
\method + KD (LB=5, MBR=10)   & -    & -    & 30.62    & 26.29 & 33.29    &  32.59   \\
\method + KD (LB=10, MBR=5)   & -    & -    & 30.76    & 26.04 & 33.40    &  32.89   \\
\bottomrule
\end{tabular}
}
\end{table}

\section{Implementation Details}
\label{sec: implementation details}
All our implementations are based on \texttt{Transforme-base}~\citep{vaswani2017attention} for all datasets except IWSLT14. 
For IWSLT14, we use a smaller architecture that has 4 attention heads and 1024-dimensional feedforward layers. 
The embedding dimension for the diffusion model is 16 on IWSLT14 and 64 on the others. 
In the implementation of our method, we follow recent advances and apply self-conditioning techniques~\citep{cdcd,chen2022analog,strudel2022self}. Besides, following previous practice in non-autoregressive machine translation, we train our model both with and without knowledge distillation~\citep[KD,][]{kim2016sequence,zhou2020understanding}.

During inference, we apply beam search in the autoregressive Transformer with beam size~5 for machine translation.
Correspondingly, we use length beam 5 in the non-autoregressive models, except for CDCD and DiffuSeq since they vary the target lengths by predicting paddings instead of length predictions.
For text simplification and paraphrasing, we report results with various length beams as length prediction on these tasks is more challenging and less studied. 
For all the diffusion-based methods, we follow previous work~\citep{diffusionlm,gong2022diffuseq,cdcd} and apply Minimum Bayes-Risk (MBR) decoding~\citep{kumar2004minimum}.
For both DiffusionLM and our model, we sample with 20 steps.

We implement DiffusionLM and \method upon \code{fairseq}~\citep{ott2019fairseq}, and also train Transformer and CMLM baselines using \code{fairseq}.
For data preprocessing, we follow the instruction in \code{fairseq} for IWSLT14\footnote{\url{https://github.com/facebookresearch/fairseq/tree/main/examples/translation}} and use the preprocessed data by~\cite{gu2021fully} for WMT14 and WMT16\footnote{\url{https://github.com/shawnkx/Fully-NAT}}. 
For Wiki and QQP, we use preprocessed data provided by DiffuSeq\footnote{\url{https://github.com/Shark-NLP/DiffuSeq}} and tokenized them with byte-pair encoding \citep[BPE,][]{bpe}.  
The training batch size is 128K for WMT14/WMT16, and 32K for the others.
We empirically find checkpoint averaging unnecessary for our method and have not applied it in all our implementations.

\section{Relationship Between Different Noise Schedules and Time Samplers}
\label{sec:appendix scheduler}

Generally, the training objective of diffusion models can be expressed as 
$$
\mathbb{E}_{t\sim r(t), \bm{\epsilon}\sim\mathcal{N}(\bm{0}, \bm{I})}[w(t)\|\bm{z}_{\bm{\theta}}-\bm{z}(0)\|_2^2],
$$
where $\bm{z}_{\bm{\theta}}$ is the model prediction $\bm{z}_{\bm{\theta}}(\sqrt{1-\sigma^2(t)}\bm{z}(0)+\sigma(t)\bm{\epsilon}, t)$ for short.

The above expectation over timesteps can be rewritten into the expectation of noise scales as follows.
$$
\begin{aligned}
    &\mathbb{E}_{t\sim r(t), \bm{\epsilon}}[w(t)\|\bm{z}_{\bm{\theta}}-\bm{z}(0)\|_2^2]\\
    =&\mathbb{E}_{\bm{\epsilon}}[\int_0^1 r(t)w(t)\|\bm{z}_{\bm{\theta}}-\bm{z}(0)\|_2^2\mathrm{d}t]\\
    =&\mathbb{E}_{\bm{\epsilon}} [\int_0^1 \hat r(\sigma) \hat w(\sigma) \|\bm{z}_{\bm{\theta}}-\bm{z}(0)\|_2^2] \frac{\mathrm{d}t}{\mathrm{d}\sigma}\mathrm{d}\sigma\\
    =&\mathbb{E}_{\sigma\sim U(0,1), \bm{\epsilon}}[w^\prime(\sigma) \|\bm{z}_{\bm{\theta}}-\bm{z}(0)\|_2^2],
\end{aligned}
$$ 
where $w^\prime(\sigma)=w(\sigma^{-1})r(\sigma^{-1})\frac{\mathrm{d}t}{\mathrm{d}\sigma}$. Therefore, training with different noise schedules and different time samplers is interchangeable by applying different weighting functions.

\section{Effect of Sizes of Length Beam and MBR}
\label{sec: lb and mbr}
\begin{figure}[h]
    \centering
    \includegraphics[width=0.5\linewidth]{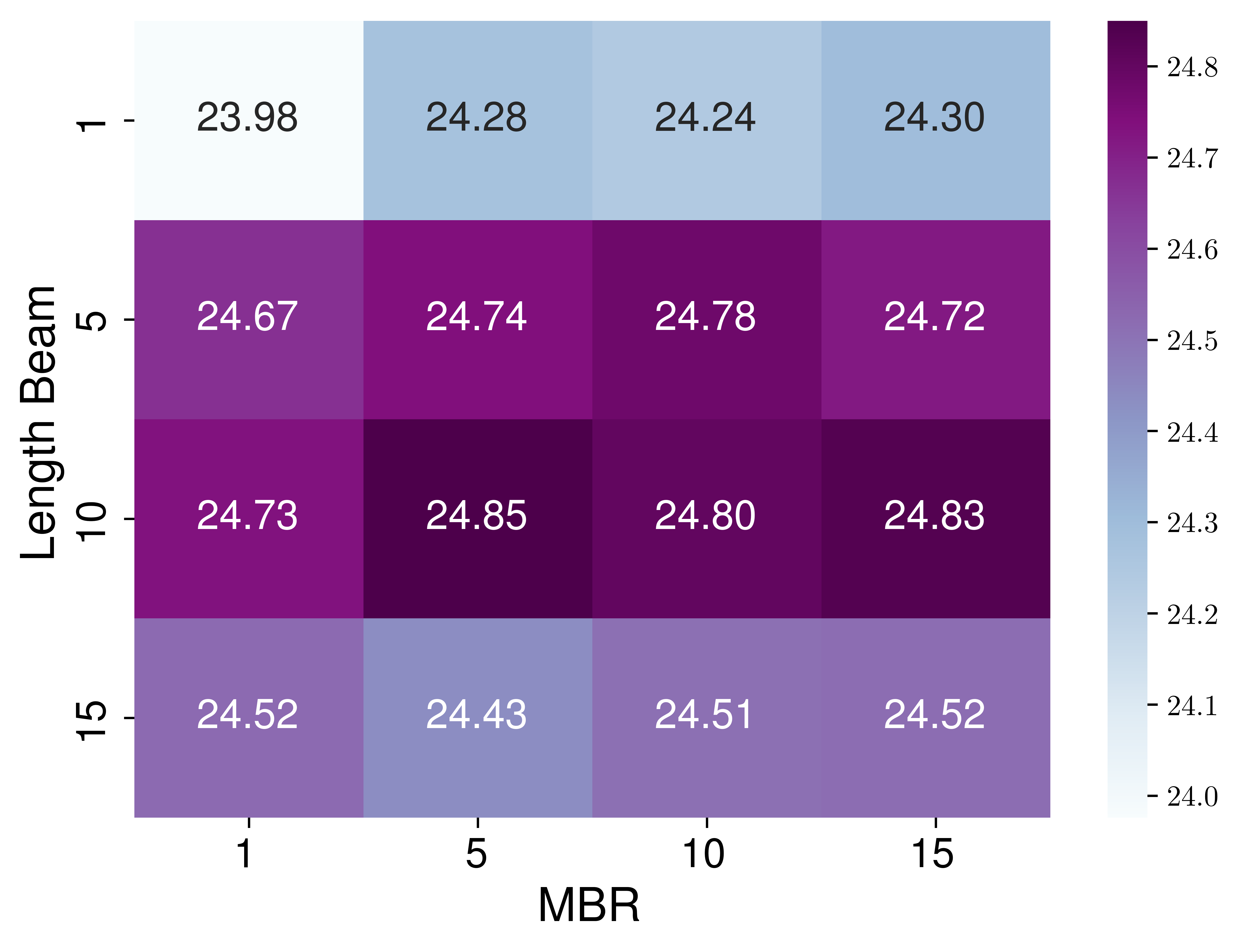}
    \caption{SacreBLEU on the validation set of WMT14 \textsc{En}$\rightarrow$\textsc{De} with different length beams and MBR sizes.}
    \label{fig:lb_mbr}
\end{figure}

\method can leverage both length beam search and MBR decoding to produce diverse candidates for selection.
The results on all of the evaluated datasets (Tab.~\ref{tab:mt-result} and Tab.~\ref{tab:diffuseq-data}) demonstrate that the method can gain its performance by properly adjusting the two hyperparameters for sampling.
In particular, we search on various combinations of length beams and MBRs and evaluate the corresponding performance of \method on the validation set of WMT14 \textsc{En}$\rightarrow$\textsc{De}, shown in Fig.~\ref{fig:lb_mbr}. 
The model performance rises first and then drops down as the length beam increases. 
And for each length beam, we can further boost the performance of \method with MBR $> 1$, suggesting that the effects of the two factors are complementary.

Using both length beam search and MBR decoding also brings benefits to \method over those only involving one of them.
Compared to CMLM which decodes deterministically, \method is able to sample multiple sentences for each length beam, providing more diverse candidates. 
Compared to CDCD, which predicts paddings to generate sentences of various lengths and whose sampling efficiency is restricted by maximum target length, \method's use of length beams allows more fine-grained control of the computational budget.

\end{document}